\documentclass{article}

\usepackage[preprint]{neurips_2025}
\usepackage{multirow}
\usepackage{amsmath}
\usepackage{float}
\usepackage{graphicx}
\usepackage{wrapfig}
\usepackage{algorithm}
\usepackage{algorithmic}

\usepackage[utf8]{inputenc} 
\usepackage[T1]{fontenc}    
\usepackage{hyperref}       
\usepackage{url}            
\usepackage{booktabs}       
\usepackage{amsfonts}       
\usepackage{nicefrac}       
\usepackage{microtype}      
\usepackage{xcolor}         

\title{Any-to-Any Learning in Computational Pathology via Triplet Multimodal Pretraining}

\makeatletter
\def\thanks#1{\protected@xdef\@thanks{\@thanks
        \protect\footnotetext{#1}}}
\makeatother

\author{
Qichen Sun$^{\dagger 1}$,\; Zhengrui Guo$^{\dagger 2}$,\; Rui Peng$^{1}$,\; Hao Chen$^{2}$,\; Jinzhuo Wang$^{\ast 1}$ \thanks{$\ast$ Corresponding authors, $\dagger$ equal contribution.}
\\
 $^{1}$Peking University,\; $^{2}$The Hong Kong University of Science and Technology\\
\tt \small qc\_sun@stu.pku.edu.cn,~wangjinzhuo@pku.edu.cn
\vspace{-1em}
}

\begin{document}

\maketitle

\begin{abstract}
  Recent advances in computational pathology and artificial intelligence have significantly enhanced the utilization of gigapixel whole-slide images  and and additional modalities (e.g., genomics) for pathological diagnosis. Although deep learning has demonstrated strong potential in pathology, several key challenges persist: (1) fusing heterogeneous data types requires sophisticated strategies beyond simple concatenation due to high computational costs; (2) common scenarios of missing modalities necessitate flexible strategies that allow the model to learn robustly in the absence of certain modalities;  (3) the downstream tasks in CPath are diverse, ranging from unimodal to multimodal, cnecessitating a unified model capable of handling all modalities.
To address these challenges, we propose ALTER, an any-to-any tri-modal   pretraining framework that integrates WSIs, genomics, and pathology reports. The term "any" emphasizes ALTER’s modality-adaptive design, enabling flexible pretraining with any subset of modalities, and its capacity to learn robust, cross-modal representations beyond WSI-centric approaches.
 We evaluate ALTER across extensive clinical tasks including survival prediction, cancer subtyping, gene mutation prediction, and report generation, achieving superior or comparable performance to state-of-the-art baselines.
\end{abstract}

\section{Introduction}

Pathology, a cornerstone of clinical medicine, continues to rely on expert examination of pathological slides as the gold standard for diagnosis. Recently, deep learning techniques, particularly Foundation Models (FMs) \cite{ma2024towards,10656058,xu2024multimodal}, have been introduced to assist the diagnostic process in computational pathology (CPath). While FMs have demonstrated remarkable generalization across various domains, their application in pathology—especially on gigapixel whole-slide images (WSIs)—faces several critical challenges.

A primary barrier to progress in CPath is the  absence of effective multimodal pretraining, where fusing heterogeneous modalities demands strategies beyond simple concatenation due to computational constraints. Genomics offer insights into the molecular mechanisms underlying visual features, while pathology reports provide global contextual information that complements local tissue analyses. This underscores the importance of incorporating multimodal inputs during pretraining to facilitate the learning of transferable representations. However, multimodal learning is significantly more computationally intensive than unimodal approaches, especially given the size of WSIs. Some works mitigate this with attention approximation or cross-attention pooling \cite{10657501,9710773}, but often at the cost of model expressiveness. Consequently, many FMs focus solely on extracting slide-level features from WSIs \cite{chen2024towards, xu2024multimodal}, which limits their flexible application in a wide range of multimodal scenarios.

A further obstacle is the scarcity of fully paired multimodal datasets in CPath \cite{hemker2023healnet}. Omics profiles may be missing for certain patients, while free-text reports can vary in completeness and availability. Therefore, assuming complete modality access is often unrealistic in clinical settings.  This fragmented data landscape calls for a flexible pretraining paradigm that can effectively learn from partially observed modalities without requiring complete alignment across samples. Moreover, the downstream tasks in CPath are diverse, ranging from unimodal to multimodal, and often demand complementary information across modalities. As a result,  developing universal models which can perform modality alignment in the presence of missing modalities is an imperative issue for real-world clinical settings. 

To address these challenges, we propose \textbf{ALTER}, the \textbf{A}ny-to-any \textbf{L}earning in compu\textbf{T}ational pathology via tripl\textbf{E}t multimodal p\textbf{R}etraining, a pretraining paradigm designed to develop FMs that flexibly integrate multimodal data across diverse scenarios.  ALTER enables pretraining with any subset of three modalities—whole slide images, genomic profiles, and diagnostic reports—allowing the model to both accept arbitrary modality combinations and learn mutual cross-modal mappings for any downstream tasks. The key idea is to leverage weak supervision via contrastive learning and masked language modeling (MLM) to build robust intra-modal, inter-modal, and intra-sample constraints. These three-tiered constraints comprehensively encompass the interaction scope in multimodal fusion, supporting flexible modality combinations at both training and inference.  To reduce computational overhead, ALTER incorporates modality-specific aggregation modules for WSIs and omics, enabling efficient cross-modal interaction beyond simple concatenation.
Notably, ALTER is modular and generalizable: it  can be extended to any multimodal scenario and is compatible with different backbone FMs.
In this paper, we demonstrate the multimodal adaptability of ALTER by training an any-to-any model with 6,850 tri-modal pairs on 29 cancer sites  from The Cancer Genome Atlas (TCGA) and conduct diverse downstream task validations on 10 public datasets.

To summarize, our contributions include (1) proposing a novel pretraining paradigm  using WSIs, genomics and reports which can handle computational and missing modality problems, promote the flexible application of FMs in computational pathology; (2) developing a  modality-flexible multimodal Transformer, training an any-to-any model with 6,850 WSI-omic-report pairs on 29 cancer sites of TCGA; and (3) conducting four categories of downstream tasks ranging from cancer subtyping, survival prediction, gene expression prediction, to diagnostic report generation on 10 public datasets, showing the superior performance and reliable generalization of ALTER.

\section{Related Work}
\label{rw}

 \noindent\textbf{Supervised Learning in CPath.} 
  One prominent approach in supervised CPath is \textit{multiple instance learning} (MIL). WSIs present significant computational challenges due to their massive size and GPU memory limitations \cite{araujo2019computing}. This computational constraint, coupled with the scarcity of fine-grained annotations, has led to the emergence of weakly supervised learning approaches in computational pathology, where only slide-level labels are available \cite{van2021deep}. MIL is a representative class in this domain, operating through three essential steps: (1) WSI preprocessing, which includes tissue segmentation and patch tiling; (2) feature extraction using pre-trained deep learning models; and (3) patch feature aggregation to generate slide-level representations \cite{ilse2018attention,campanella2019clinical,chikontwe2020multiple,lu2021data,shao2021transmil,li2021dt,xiang2022dsnet,hou2022h,zheng2022graph,zhang2022dtfd,wang2022scl,yu2023prototypical,lu2023visual,lin2023interventional,li2024dynamic,yang2024mambamil}. Research efforts in MIL primarily focus on developing methods to better capture the complex correlations among numerous WSI patches, aiming to generate more discriminative slide-level representations and improve diagnosis performance. 

 Beyond vision-only modeling, there is growing interest in \textit{multi-modal learning},  where WSIs are integrated with additional modalities such as genomics or clinical texts \cite{boehm2022harnessing}. A prominent direction in this field is the fusion of genomic data with pathological features for survival prediction (prognostication), which leverages both the morphological patterns visible in WSIs and the underlying molecular characteristics of the disease
 \cite{ash2021joint,chen2020pathomic,chen2022pan,jaume2024modeling,li2022hfbsurv,mobadersany2018predicting,schmauch2020deep,xie2024spatially,xu2023multimodal,zhou2023cross,song2024multimodal,meng2024genomics,hemker2023healnet}.  Several fusion mechanisms have been proposed, such as the late \cite{chen2022pan, hemker2024healnet} or early fusion \cite{xu2023multimodal, zhou2023cross}. Another line of work also focuses on aligning WSIs with textual descriptions, including patch-level captions \cite{lu2023visual} and slide-level diagnostic reports 
 \cite{guo2024histgen}.

 \noindent\textbf{Self-supervised Pretraining in CPath.} In parallel, self-supervised learning  has emerged as a powerful approach for pretraining models without relying on costly annotations.  A notable strand of research is \textit{vision-only pretraining} approaches, which leverage the inherent structural and contextual information within WSIs to create pretext tasks for model pretraining \cite{chen2022scaling,lazard2023giga,yu2023slpd,wu2023position,jiang2024masked,song2024morphological, yu2023prototypical}.  For example, Chen \textit{et al.} \cite{chen2022scaling} propose a hierarchical three-stage pretraining pipeline that progressively aggregates representations from patch-level to region-level, and ultimately to slide-level.

Recently, \textit{multi-modal pretraing} approaches have also gained momentum.
Recent advances in self-supervised learning for computational pathology have expanded to multi-modal pretraining strategies that leverage either the combination of WSIs and genomic data \cite{ding2023pathology,jin2023gene,jaume2024transcriptomics}, or multiple staining modalities of WSIs \cite{jaume2024multistain}. The first category aims to learn comprehensive representations by jointly capturing morphological patterns from histological images and their corresponding molecular characteristics in an unsupervised manner. For instance, TANGLE \cite{jaume2024transcriptomics} employs contrastive learning to align detailed molecular information from expression profiles with the visual features of the tissue.  The second category, represented by \cite{jaume2024multistain}, treats WSIs stained with multiple biomarkers as different augmentation views and learns robust slide representations via global-local cross-stain contrastive learning objectives.

 \noindent\textbf{Foundation Models in CPath.} The emergence of foundation models has marked a significant advancement in computational pathology, with numerous architectures being developed to enhance diagnostic and prognostic capabilities through large-scale pathological data pretraining \cite{ma2024towards,chen2024towards,xu2024whole,vorontsov2024foundation,wang2024pathology,lu2024visual,huang2023visual,xu2024multimodal}. 
 These models can be broadly categorized into two groups: unimodal architectures that focus exclusively on visual information from WSIs,  such as \cite{ma2024towards,chen2024towards,xu2024whole,vorontsov2024foundation,wang2024pathology}, as well as multimodal architectures that combine WSIs with additional data sources including pathology reports or genomic expression profiles, as demonstrated by CONCH \cite{lu2024visual,huang2023visual,xu2024multimodal}.
 Among these works, while mSTAR \cite{xu2024multimodal} represents the first attempt to jointly utilize WSIs, genomic data, and pathology reports, it primarily functions as a visual feature extractor and lacks the flexibility to process and integrate arbitrary modalities. In contrast, our work fundamentally addresses this limitation by enabling seamless integration and processing modalities in an any-to-any manner.

\section{Method}
\label{meth}

\subsection{Problem Formulation}

Given a triplet $(W, G, R)$—where $W$ denotes a WSI, $G$ for the corresponding gene expressions, and $R$ for the diagnostic report—we first decompose $W$ into non-overlapping patches $\{p_{1}, p_{2}, \dots, p_{N_h}\}$ and tokenize $R$ into a sequence $T \in \mathbb{R}^{N_{t}}$, with $G \in \mathbb{R}^{N_{g}}$. Each patch is then processed using a pretrained feature extractor $\Theta$. In this study, we adopt UNI \cite{chen2024towards}, a foundation model for computational pathology, as the extractor. Formally, we obtain patch features $h_i = \Theta(p_i) \in \mathbb{R}^{d_{w}}$, forming the feature bag $H = \{h_{1}, h_{2}, \dots, h_{N_h}\}$. As a result, the original triplet is converted into a serialized input $(H, G, T)$. Our objective is to learn a robust mapping $f: \mathcal{X} \rightarrow \mathcal{Y}$ in a weakly supervised manner, where $\mathcal{X}$ denotes the space of serialized triplets and $\mathcal{Y}$ is the latent representation space of the three modalities. The central challenge is to align the modalities effectively, ensuring that the latent space captures both modality-specific characteristics and cross-modal correlations.

 \begin{figure*}[t]
\centering
\includegraphics[width=\textwidth]{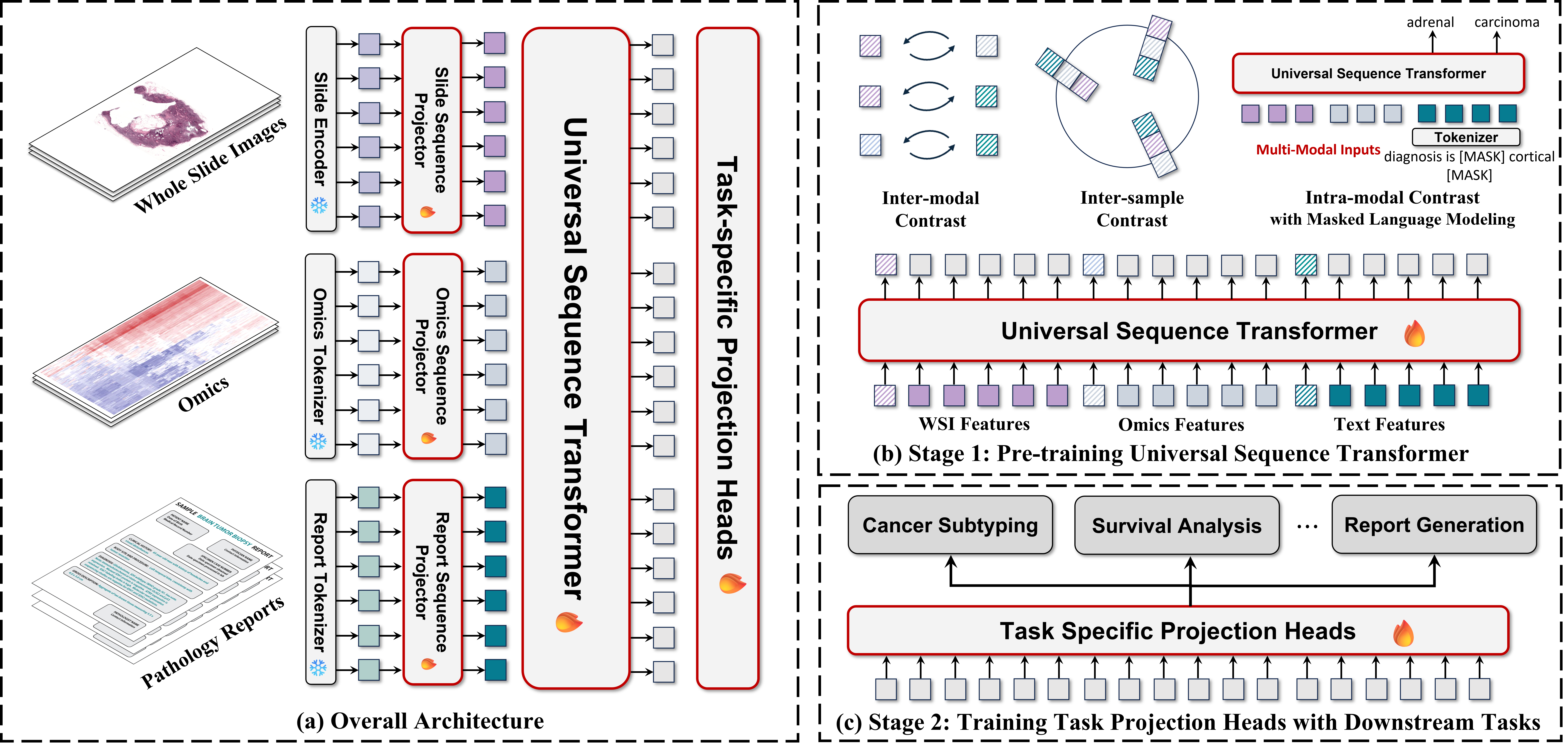}
\caption{Overview of our pretraining framework, \textbf{ALTER}. (a) ALTER processes each modality using modality-specific encoders, followed by a universal sequence Transformer and task-specific projection heads for downstream prediction. (b)  The three-tiered constraints of ALTER, which can enable model to align multimodal inputs without requiring full modality pairing. (c) ALTER can be applied to any downstream task by integrating task-specific projection heads for fine-tuning. } \label{fig:method}
\vspace{-1.2em}
\end{figure*}

\subsection{Modality Encoder}
\label{me}
As shown in Fig.~\ref{fig:method} (a), we use modality-specific encoders in the early stages of the model, which integrate intramodal information and prepare for subsequent fusion. 

Due to high input lengths, $N_h \approx  20{,}000$, $N_g \approx 10{,}000$, and $N_t \approx 500$, direct multimodal fusion is computationally intensive. The concatenated sequence $\mathrm{Y} \in \mathbb{R}^{(N_h + N_g + N_t) \times d}$ incurs $\mathcal{O}((N_h + N_g + N_t)^2)$ memory under self-attention \cite{vaswani2017attention}:
\begin{equation}
    \mathrm{Y_{att}} = \mathrm{softmax}\left( \frac{\mathrm{Y}W_q (\mathrm{Y}W_k)^T}{\sqrt{d}} \right) \mathrm{Y}W_v.
\end{equation}

Thus, feature aggregation, especially for WSIs and RNA-Seq, is essential during encoding. In this stage, we propose modality-specific aggregation algorithms tailored to the characteristics of the above two  modalities, aiming to capture more holistic features while reducing computational overhead.

\textbf{Whole Slide Images.}  We adopt a 2-layer TransMIL \cite{shao2021transmil} as the slide encoder to project WSI features into a $d$-dimensional space. The process can be formulated as follows:
\begin{equation}
    H^{(1)} = \mathrm{TransMIL} ( \{h_{1}, h_{2}, \dots, h_{N_h} \} )\\
        = \{ h^{(1)}, h^{(1)}_{1}, h^{(1)}_{2}, \dots, h^{(1)}_{N_h} \} \in \mathbb{R}^{(N_h + 1) \times d},
\end{equation}
where $H = \{h_{1}, h_{2}, \dots, h_{N_h} \} \in \mathbf{R}^{N_h \times d}$ is the slide features bag after sampling, and $h^{(1)}$ is the $d$-dimensional [CLS] token.

To reduce the computational cost of subsequent fusion, we further aggregate the obtained features after obtaining the [CLS] token and re-embedding of WSIs. We denote $\tilde{H}^{(1)} = \{h^{(1)}_{1}, h^{(1)}_{2}, \dots, h^{(1)}_{N_h} \}$, which is the re-embedding obtained through slide encoder.  To model the corresponding spatial relationships between patches, we reshape $\tilde{H}^{(1)}$ to the 2-D feature map:
\begin{equation}
    \tilde{H}^{(1)}: \mathbb{R}^{N_h \times d} \to \mathbb{R}^{([\sqrt{N_h}] \times [\sqrt{N_h}] ) \times d}.
\end{equation}
Although tumor slices exhibit high heterogeneity, the continuity of adjacent regions and the redundancy of slide-level information allow us to further compress the sequence length of pathological features. Considering the continuity  of adjacent patches, we divide the 2-D feature map into non-overlapping regions of size $a\times b$, and then we replace the features of each region with the mean of that region to obtain 
aggregated feature $\tilde{H}^{(2)}$. In the end, we reshape $\tilde{H}^{(2)}$ to a 1-D feature and concatenate it with the [CLS] token:
\begin{equation}
\begin{split}
    \tilde{H}^{(2)}&: \mathbb{R}^{([\sqrt{N_h}/a] \times [\sqrt{N_h}/b] ) \times d} \to \mathbb{R}^{\lfloor \frac{N_h}{ab} \rfloor \times d}, \\
    H^{(3)}&= \mathrm{Cat}(h^{(1)}, \tilde{H}^{(2)}) \in \mathbb{R}^{(\lfloor \frac{N_h}{ab}  \rfloor + 1) \times d}.
\end{split}
\end{equation}
 This region-wise aggregation substantially reduces sequence length while preserving contextual information within the WSI, facilitating more efficient processing and enabling future multimodal interaction.

\textbf{Genomic Profiles.}  Following scBERT \cite{yang2022scbert}, we pretrain a Performer as the gene encoder. We utilize the term-frequency-analysis method to discretize gene expression values before inputting them into the gene encoder, such as:
\begin{equation}
    G^{(1)} = \mathrm{Performer}(G) 
 = \{ g^{(1)}, g^{(1)}_{1}, g^{(1)}_{2}, \dots, g^{(1)}_{N_g} \},
\end{equation}
where $G^{(1)}\in \mathbb{R}^{(N_g + 1) \times d}$ and $g^{(1)}$ is the [CLS] token for genes.

Unlike WSIs that have similarities with neighboring areas, genomic data cannot be aggregated in a similar way. Therefore, we adopt pathways to model the interactions between genes for aggregation following previous works \cite{song2024multimodal}. A pathway refers to a group of genes involved in a specific biological process. By matching with existing biological pathway databases, we group genes into pathways, ultimately resulting in $N_p$ pathways. Finally, each pathway is pooled to obtain the aggregated gene features, such that:
\begin{equation}
    \begin{split}
        \tilde{G}^{(2)}&= \{ p_{1}, p_{2}, \dots, p_{N_p} \} \in \mathbb{R}^{N_{p} \times d},\\
        G^{(3)} &= \mathrm{Cat}(g^{(1)}, \tilde{G}^{(2)}) \in \mathbb{R}^{(N_{p} + 1) \times d},
    \end{split}
\end{equation}
where $p_j$ is the feature of each pathway. This pathway-level aggregation enables the model to consider genes as part of a coordinated physiological process rather than isolated units.

\textbf{Diagnostic Reports.}  There have been many efforts dedicated to providing reliable embedding for medical texts. We choose BioBERT \cite{lee2020biobert} as text encoder in this study, such that,
\begin{equation}
    T^{(1)} = \mathrm{BioBERT}(T) \in \mathbb{R}^{N_t \times d},
\end{equation}
where $N_t = 512$, and $T^{(1)}[0]$ is the [CLS] token for reports, denoted as $t^{(1)}$.

\subsection{Fusion Architecture}

 \begin{wrapfigure}{r}{0.4\textwidth}
  \centering
  \vspace{-3em}
  \includegraphics[width=0.4\textwidth]{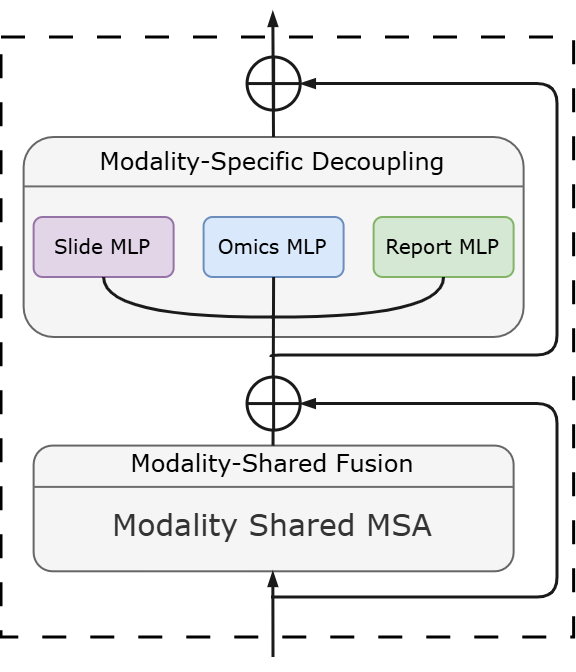}
  \caption{Universal sequence Transformer architecture of ALTER.}
  \vspace{-2em}
  \label{fig:fusion}
\end{wrapfigure}

To accommodate varying modality combinations during fusion, we design a two-stage fusion strategy within each universal sequence Transformer in Fig.~\ref{fig:method} (a): (1) modality-shared fusion; (2) modality-specific decoupling. As shown in  Fig.~\ref{fig:fusion}, we propose modality-specific mixture of experts (MoE) to  decode structural characteristics of each modality from the fused latent space. This modality MoE architecture is particularly well-suited for CPath, as it enables the model to capture high-dimensional spatial features in WSIs, sparse tabular patterns in genomics, and semantic cues in clinical text.

Specifically, we employ modality-shared self-attention layers  $\Phi(\cdot)$ to model interactions across modalities for the first stage. The shared attention matrix maps combinations of modalities into a unified latent space, $\mathrm{Z'} = \Phi( \mathrm{Z}) + \mathrm{Z}$, where $\mathrm{Z} = \{ H^{(3)}, G^{(3)}, T^{(1)} \}$ denotes the concatenation feature input and $\mathrm{Z'} =\{ H', G', T' \}$ is the resulting output. In the next stage, we introduce modality-specific experts to address the distinct statistical properties. These are defined as $ F'' = f_{F} (F') + F'$, where $F \in (H, G, T)$ represents the specific modality features, and $f_{F}$ is the corresponding expert  module. 

Notably, the two-stage design inherently supports missing modality scenarios. Since attention is computed over the available subset of modalities, the model accommodates incomplete inputs without requiring imputation or substitution—common sources of noise in traditional fusion approaches. This property not only ensures robust inference but also permits effective training on partially observed data, maximizing the utility of real-world clinical datasets, where missing modalities are frequent.

\subsection{Pretraining Task}
To fully leverage information from multiple modalities and achieve deep fusion, we design a pretraining task comprising three hierarchical levels, as illustrated in Fig.~\ref{fig:method} (b). These three-tiered constraints comprehensively capture the interaction scope of multimodal fusion, enabling the model to learn effectively under varying input conditions. This approach allows the model to be trained with arbitrary modality combinations, enhancing its generalization across diverse downstream tasks.

\textbf{Intra-modal Level.}  In this level, we focus primarily on the specific information of modalities to obtain better representations. We utilize the MLM objective on unimodal and multimodal data. For WSIs, we apply the same region grouping method as in \S~\ref{me} to the original features. We randomly select some regions to mask and use the mean of the original features within the regions as labels. For omics, we randomly mask a certain proportion of gene expression values in each pathway.  For reports, we adopt the original MLM approach. 

Although we use MLM to focus more on specific information within each modality, when multiple modalities are combined as input, it can also learn the alignment between different modalities. During the actual training process, for multi-modal inputs, after every 10 epochs, we randomly select a modality to perform the MLM task, denoted as $\mathcal{L}_{MLM}$, as follows:
\begin{equation}
    \mathcal{L}_{\text{MLM}} = -\frac{1}{N} \sum_{i=1}^{N} \sum_{j \in \mathcal{M}} \log P(x^{(i)}_{j} | x^{(i)}_{-j}),
\end{equation}
where $x^{(i)}_j$ is the $i$-th original token at position $j$, and $x^{(i)}_{-j}$ represents the context tokens excluding the masked token $x^{(i)}_j$.

\textbf{Inter-modal Level.}  At this level, we focus on aligning different modalities to inject multimodal information into our model. Given the [CLS] tokens set $(h^{(1)}, g^{(1)}, t^{(1)})$, we construct CLIP \cite{radford2021learning} loss between the three pairs to facilitate contrastive learning. Take WSI-gene pairs as example, the calculation process can be formalized as follows:
\begin{equation}
    \begin{split}
        \mathcal{L}_{H-G}=  -\frac{1}{2 N} \sum_{i=1}^{N} \log \frac{\exp \left(h_{i}^{(1)\top} g^{(1)}_{i} / \tau\right)}{\sum_{j=1}^{N} \exp \left(h_{i}^{(1)\top} g^{(1)}_{j} / \tau\right)} 
 -\frac{1}{2 N} \sum_{j=1}^{N} \log \frac{\exp \left(g_{j}^{(1)\top} h^{(1)}_{j} / \tau\right)}{\sum_{i=1}^{N} \exp \left(g^{(1)\top}_{j} h^{(1)}_{i} / \tau\right)},
    \end{split}
\end{equation}
where $\{ (h^{(1)}_i, g^{(1)}_i)\}_{i=1}^{N}$ is the WSI-gene pairs in a mini-batch, $N$ is the number of samples in a mini-batch, $\tau$ is a scale factor. Similarly, we have $\mathcal{L}_{G-T}$ and $\mathcal{L}_{T-H}$. Finally, we can get $\mathcal{L}_{CLIP}$ such as $\mathcal{L}_{CLIP} = \mathcal{L}_{H-G} + \mathcal{L}_{G-T} + \mathcal{L}_{T-H}$.

\textbf{Inter-sample Level.} Given the distinctive characteristics of pathological data, there should be significant distinctions between features of different cancers. Hence, we utilize cancer categories as sample-level labels to conduct contrastive learning at this level. Through this task, the model learns global features about cancer to better adapt to various scenarios. Specifically, we concatenate the [CLS] tokens of a sample to obtain a sample-level [CLS] token. Subsequently, based on the cancer type of the sample, we apply triplet loss to different samples, such as,
\begin{equation}
    \mathcal{L}_{triplet} = \frac{1}{N} \sum^{N} \max \left(d(a_{i}, p)-d\left(a_{i}, n\right)+\epsilon, 0\right),
\end{equation}
where $a_{i}$ is the anchor sample, $p$ is the positive sample and $n$ is the negative sample, $d(\cdot)$ is the metric function, $\epsilon$ is the margin.

The overall loss function is $\mathcal{L} = \alpha \mathcal{L}_{MLM} + \beta \mathcal{L}_{CLIP} + \mathcal{L}_{triplet}$, where $\alpha$ and $\beta$ are hyperparameters.

\begin{table*}[t]
\caption{C-index (mean) over four cancer datasets across five folds. The best results are highlighted in \textbf{bold}, second best \underline{underlined}. ALTER outperforms all of the baselines acroos four cancer types.} 
\centering
\label{surv}
\renewcommand{\arraystretch}{1.3}
\resizebox{\linewidth}{!}{
\begin{tabular}{c|c|cccccc}
\hline \hline
\textbf{Methods}   & \textbf{Modality}      & \textbf{UCEC}   & \textbf{STAD}   & \textbf{KIRP}   & \textbf{KIRC}   & \textbf{Average} \\ \hline \hline
SNN  \cite{klambauer2017self}    & g.     & 0.614 $\pm$ 0.017 & 0.527 $\pm$ 0.035 & 0.753 $\pm$ 0.043 & 0.632 $\pm$ 0.037 & 0.634 \\ \hline
TransMIL \cite{shao2021transmil}  & h.    & \underline{0.717 $\pm$ 0.022}& 0.628 $\pm$ 0.116 & 0.658 $\pm$ 0.056 & 0.677 $\pm$ 0.032 & 0.670  \\
ABMIL \cite{ilse2018attention}    & h.    & 0.710 $\pm$ 0.017 & 0.611 $\pm$ 0.093 & 0.661 $\pm$ 0.051 & 0.678 $\pm$ 0.038& 0.664  \\ \hline
MCAT \cite{9710773}      & g.+h.   & 0.707 $\pm$ 0.027 & 0.651 $\pm$ 0.060 & \underline{0.818 $\pm$ 0.047} & 0.688 $\pm$ 0.044 & 0.716  \\
Porpoise \cite{chen2022pan}   & g.+h.   & 0.704 $\pm$ 0.021 & 0.627 $\pm$ 0.088& 0.765 $\pm$ 0.058& 0.719 $\pm$ 0.043 & 0.704  \\
MOTCat \cite{xu2023multimodal}   & g.+h.     & 0.714 $\pm$ 0.057& 0.633 $\pm$ 0.104 & 0.796 $\pm$ 0.078 & 0.703 $\pm$ 0.051 & 0.711  \\ 
SurvPath \cite{10657501} & g.+h.    &  0.705 $\pm$ 0.039 & 0.636 $\pm$0.065 & 0.813 $\pm$ 0.063 & \underline{0.743 $\pm$ 0.145} & \underline{0.724} \\
PIBD  \cite{zhang2024prototypical}     & g.+h.   & 0.702 $\pm$ 0.078& \underline{0.652 $\pm$ 0.137} & 0.804 $\pm$ 0.082 & 0.730 $\pm$ 0.073 & 0.722  \\ 
\hline
ALTER(Ours)     & g.+h.     & \textbf{0.745 $\pm$ 0.049} & \textbf{0.695 $\pm$ 0.091} & \textbf{0.851 $\pm$ 0.046} & \textbf{0.754 $\pm$ 0.048} & \textbf{0.762} \\ \hline \hline
\end{tabular}}
\vspace{-1em}
\end{table*}

\subsection{Model Deployment}
As illustrated in Fig.~\ref{fig:method} (c), we append task-specific heads to ALTER for fine-tuning on downstream tasks. Notably, we do not impose any architectural constraints on the task heads. For multimodal tasks, we employ cross-attention to re-embed the CLS tokens, while for other tasks, we use  MLPs.

\begin{table*}[t]
\caption{Results of cancer subtyping on 4 datasets. The best results are highlighted in \textbf{bold}, second best \underline{underlined}.} 
\centering
\label{sub}
\renewcommand{\arraystretch}{1.3}
\resizebox{\textwidth}{2.1cm}{
\begin{tabular}{c|cccccccccccccc}
\hline \hline
\multicolumn{1}{c|}{\multirow{2}{*}{\textbf{Methods}}}      & \multicolumn{2}{c}{{\textbf{BRACS}}} & & \multicolumn{2}{c}{\textbf{PANDA}} & & \multicolumn{2}{c}{\textbf{TUPAC-16}} & & \multicolumn{2}{c}{\textbf{UBC-OCEAN}} & & \multicolumn{2}{c}{\textbf{Average}}  \\
\cline{2-3} \cline{5-6} \cline{8-9} \cline{11-12} \cline{14-15}
& AUC & F1 Score & & AUC & F1 Score & & AUC & F1 Score & & AUC & F1 Score & & AUC & F1 Score \\ \hline\hline 
Mean Pooling & 0.804 & \underline{0.446} & &  \textbf{0.951} & \textbf{0.734} & & 0.665 & 0.455 & &0.970 & 0.845 & & 0.877 & 0.620 \\
Max Pooling & 0.776 & 0.427 &  & 0.919 & 0.642 & & 0.688 & \underline{0.548} & & 0.971 & 0.813 & & 0.870 & 0.608 \\
WIKG  \cite{li2024dynamic} & \underline{0.823} & 0.429 &  & \underline{0.942} & 0.741 & & 0.630 & 0.457 & & 0.972 & \underline{0.869} &  & 0.872 & 0.624 \\
ABMIL \cite{ilse2018attention} & 0.811 & 0.442 & & 0.930 & 0.683 & & 0.702 & 0.539 & & 0.972  & 0.832 & & \underline{0.883} & \underline{0.625}\\
DS-MIL \cite{li2021dual} & 0.790 & 0.439 &  & 0.903 & 0.591 & & \underline{0.707} &0.538 & &  \textbf{0.976} & 0.830 & & 0.874 & 0.600 \\
TransMIL \cite{shao2021transmil} & 0.729 & 0.346 &  & 0.916 & 0.644 &  & 0.656 & 0.540 & & 0.966 & 0.803 & & 0.853 & 0.584 \\
CLAM-SB \cite{lu2021data} & 0.781 & 0.434 &  & 0.930 & 0.688  & & 0.687 & 0.453 & & 0.972 & 0.811 & & 0.873 & 0.597 \\ \hline
ALTER(Ours) & \textbf{0.844}  & \textbf{0.455} & & 0.932 & \underline{0.692} & & \textbf{0.758} & \textbf{0.612} & & \underline{0.973} & \textbf{0.874} & & \textbf{0.901} & \textbf{0.658} \\ \hline \hline
\end{tabular}
}
\vspace{-0.9em}
\end{table*}

\section{Experiments}
\label{exp}

\begin{table}[t]
\caption{Results of gene mutation prediction on two datasets. ALTER outperforms all of the baselines.} 
\centering
\vspace{-0.5em}
\label{gene}
\renewcommand{\arraystretch}{1.3}
\resizebox{\textwidth}{!}{
\begin{tabular}{c|cccccccc}
\hline \hline
\textbf{Methods}        & Mean Pooling & Max Pooling & WIKG & ABMIL & TransMIL & DS-MIL & CLAM-SB & ALTER(Ours) \\ \hline\hline 
TP53          & 0.764        & \underline{0.801} & 0.794 & 0.747 & 0.707   & 0.786  & 0.747   & \textbf{0.809} \\ 
EGFR         & 0.796        & 0.699            & 0.763 & 0.805 & 0.697   & 0.732  & \underline{0.807} & \textbf{0.811} \\ 
\hline
Average       & \underline{0.780}        & 0.750            & 0.778 & 0.776 & 0.702   & 0.759  & 0.777 & \textbf{0.810} \\ \hline \hline
\end{tabular}
}
\vspace{-1.7em}
\end{table}

\subsection{Datasets and Tasks}
We use 6,850 WSI-omic-report pairs  from TCGA as our pretraining dataset. To prevent data leakage, we selected four cancer types for validation in downstream tasks, while all data corresponding to the remaining 29 cancer types were included in the pretraining.

We evaluate our pretraining framework across 4 types of tasks over 10 public datasets. Since WSIs are often the most critical modality in computational pathology, we carefully selected the following four downstream tasks, each focusing on multimodal alignment, and representation of WSI-to-itself, WSI-to-Genomics, WSI-to-Report.

\textbf{(1) Survival prediction (multimodality).} Four subsets of
TCGA are used to build a survival prediction task: Uterine Corpus Endometrial Carcinoma (UCEC) ($n = 480$), Stomach Adenocarcinoma (STAD)  ($n=317$), Cervical Kidney Renal Papillary Cell Carcinoma (KIRP) 
 ($n=284$), and Kidney Renal Clear Cell Carcinoma (KIRC) ($n = 218$).

\textbf{(2) Cancer subtyping (\boldmath{$h.\rightarrow h.$}).} Four public datasets are used to predict cancer subtypes. BRACS \cite{brancati2022bracs} ($n=547$), a dataset  contains 6 different subtypes of lesions. PANDA \cite{bulten2022artificial} ($n = 10,202$),  a dataset includes six categories of prostate cancer samples. TUPAC-16 \cite{veta2019predicting} ($n=821$), a dataset includes whole slide images with known tumor proliferation scores. UBC-OCEAN \cite{asadi2024machine} ($n = 527$), an ovarian cancer dataset contains 5 cancer subtypes. 

\textbf{(3) Gene mutation prediction (\boldmath{$h. \rightarrow g.$}).}  We use a subset, lung adenocarcinoma (LUAD) (n = 412), of TCGA to predict TP53 and EGFR gene
mutation from lung adenocarcinoma WSIs.

\textbf{(4) Report generation (\boldmath{$h. \rightarrow t.$}).} We employ PatchGastricADC22 \cite{tsuneki2022inferencecaptionshistopathologicalpatches} ($n = 991$) for report generaton task, which is a large dataset consisting of 262,777 patches extracted from 991 WSIs.

\subsection{Baselines}
\label{baseline}
We employ the current SOTA models, which are mainly categorized into three types: multimodal methods, unimodal classification methods, and unimodal generation methods. For pathology, all models use the same pretrained feature extractor as ALTER based on UNI \cite{chen2024towards}.

\textbf{Multimodal methods.} We compare with \textsuperscript{\dag}\textit{MCAT} \cite{9710773}, which uses genomic-guided cross attention, \textit{Porpoise} \cite{chen2022pan}, which uses modality-specific self-attention blocks, \textit{MOTCat} \cite{xu2023multimodal}, which uses Optimal Transport, \textit{SurvPath} \cite{10657501}, which uses pathways to organize genomic data, and \textit{PIBD} \cite{zhang2024prototypical}, which fits the multimodal data distribution based on information bottleneck theory.

\textbf{Unimodal classification methods.} For histology, we employ baselines including \textit{Mean/Max Pooling}, \textit{WIKG} 
 \cite{li2024dynamic}, \textit{ABMIL} \cite{ilse2018attention}, \textit{DS-MIL} \cite{li2021dual},\textsuperscript{\dag}\textit{TransMIL} \cite{shao2021transmil}, \textit{CLAM-SB} \cite{lu2021data}. For the genomic data, we compare with pathway-specific \textit{SNN} \cite{klambauer2017self}.

\textbf{Unimodal generation methods.} We adapt generation methods including \textsuperscript{\dag}\textit{WSICaption} \cite{chen2024wsicaption}, which is a transformer-based model focus on report generation, and \textit{HistGen} \cite{guo2024histgen}, which uses local-global hierarchical encoder to achieve  feature aggregation.

\subsection{Implementation Details}
During the pretraining phase, we use five A100 GPUs for parallel training, setting the batch size to 60, with Adam as the optimizer. Following previous works \cite{10657501}, we perform 5-fold cross-validation for survival analysis, while for other downstream tasks, we use a 7:2:1 ratio to split the dataset into training, validation, and test sets. For all models, the data from each modality is standardized uniformly and  all of the baselines are trained with the code reported in the respective papers.

\section{Results}
\label{res}

\begin{wrapfigure}{r}{0.4\textwidth}
  \centering
  \vspace{-4em}
  \includegraphics[width=0.4\textwidth]{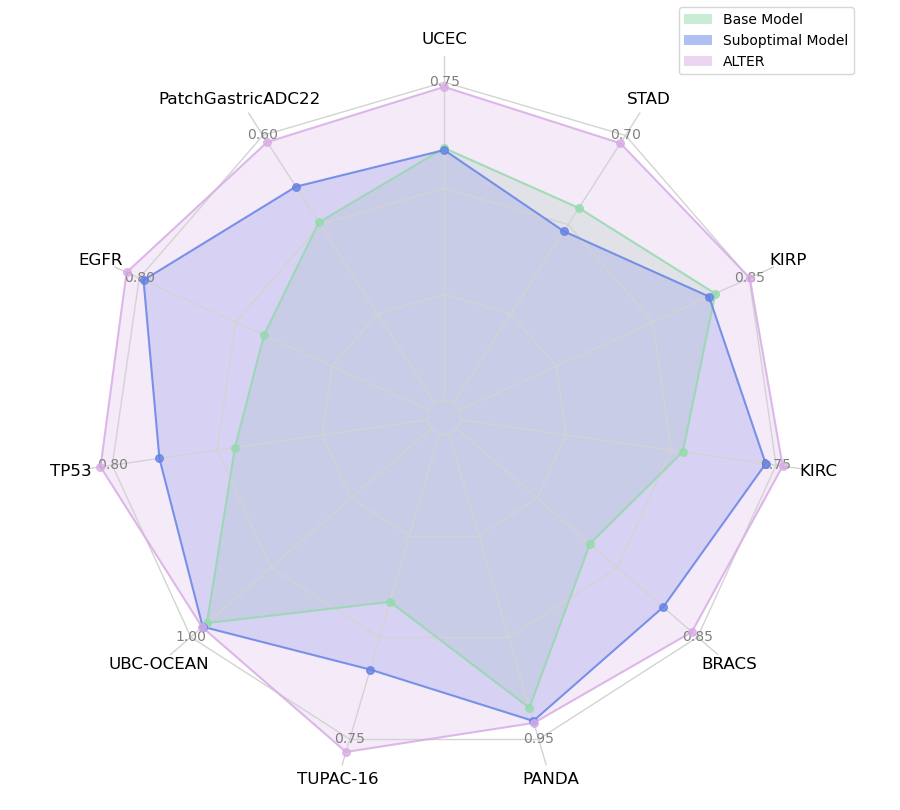}
  \caption{ALTER's overall performance across all datasets. The base model denotes commonly used task-specific benchmarks (marked with \dag~ in \S~\ref{baseline}), while the suboptimal model refers to the model with the second-best overall performance for the specific task type.}
  \vspace{-1em}
  \label{fig:radar}
\end{wrapfigure} 

As shown in Fig.~\ref{fig:radar}, we provide an overview of ALTER's performance across all tasks. It is evident that ALTER consistently achieves superior performance improvements across four types of tasks, confirming the robustness of our approach.

We present the performance of survival prediction on four cancer datasets, assessed using the C-index as shown in Tab.~\ref{surv}. Across all four cancer types, our method consistently achieves the highest performance. Compared to unimodal methods, multimodal approaches demonstrate a higher overall C-index across the four cancer types. Moreover, ALTER's overall C-index surpasses that of the second-best method by $4.1\%$, highlighting the unique potential of multimodal pretraining foundation models in CPath. 
Notably, in cases where unimodal models using different modalities exhibit large performance discrepancies (e.g., UCEC), most multimodal models fail to show significant gains, likely due to modality-specific noise. In contrast, ALTER demonstrates notable improvements, underscoring its robustness to noisy modality inputs.

Moreover, Tab.~\ref{sub} shows the subtype classification results across four public datasets.  Note that among all the methods, the proposed ALTER achieves superior performance in 3 out of 4 benchmarks and outperforms the second-best method by $2.7\%$ in overall AUC, and $5.3\%$ in overall F1 Score. The performance in the subtype classification task further indicates that the information gain from multimodal pretraining is generalizable, as the foundation model demonstrates excellent performance even in simple unimodal classification.

For gene mutation prediction tasks, as we can see in Tab.~\ref{gene}, our method demonstrates superior performance with an AUC of $0.809$ for TP53, and $0.811$ for EGFR. This indicates that our proposed framework can learn information from different modalities during the pretraining phase, even when only one modality is involved during fine-tuning.

For WSI report generation task, our experimental results demonstrate significant improvements over state-of-the-art methods. As shown in Tab.~\ref{exp:report}, our proposed approach consistently outperforms existing methods across all evaluation metrics. Specifically, our model achieves BLEU-1, BLEU-2, BLEU-3, and BLEU-4 scores of 0.628, 0.547, 0.491, and 0.450 respectively, surpassing both WSICaption and HistGen baselines by notable margins. These comprehensive improvements across all metrics demonstrate the robustness and effectiveness of our proposed method.

\begin{table}[t]
\caption{Results of report generation with several commonly used natural language generation metrics.} 
\centering
\vspace{-0.5em}
\label{exp:report}
\renewcommand{\arraystretch}{1.3}
\resizebox{0.75\linewidth}{!}{
\begin{tabular}{c|cccccc}
\hline \hline
\textbf{Methods}         & \multicolumn{1}{c}{\begin{tabular}[c]{@{}c@{}}BLEU$_{1}$\end{tabular}} & \multicolumn{1}{c}{\begin{tabular}[c]{@{}c@{}}BLEU$_{2}$\end{tabular}} & \multicolumn{1}{c}{\begin{tabular}[c]{@{}c@{}}BLEU$_{3}$\end{tabular}} & \multicolumn{1}{c}{\begin{tabular}[c]{@{}c@{}}BLEU$_{4}$\end{tabular}} & \multicolumn{1}{c}{\begin{tabular}[c]{@{}c@{}}METEOR\end{tabular}} & \multicolumn{1}{c}{\begin{tabular}[c]{@{}c@{}}ROUGE-L\end{tabular}} \\ \hline\hline 
WSICaption \cite{chen2024wsicaption} & 0.548 & 0.457 & 0.393 & 0.339 & 0.285 & 0.561 \\ 
HistGen \cite{guo2024histgen} & \underline{0.615} & \underline{0.518} & \underline{0.452} & \underline{0.402} & \underline{0.311} & \underline{0.577} \\ 
\hline
ALTER(Ours) & \textbf{0.628} & \textbf{0.547}  & \textbf{0.491}  & \textbf{0.450}  & \textbf{0.325} & \textbf{0.597} \\ \hline \hline
\end{tabular}}
\vspace{-1.5em}
\end{table}

\section{Discussion}
\label{dis}
\begin{figure*}[t]
\vspace{-1em}
\centering
\includegraphics[width=0.97\textwidth]{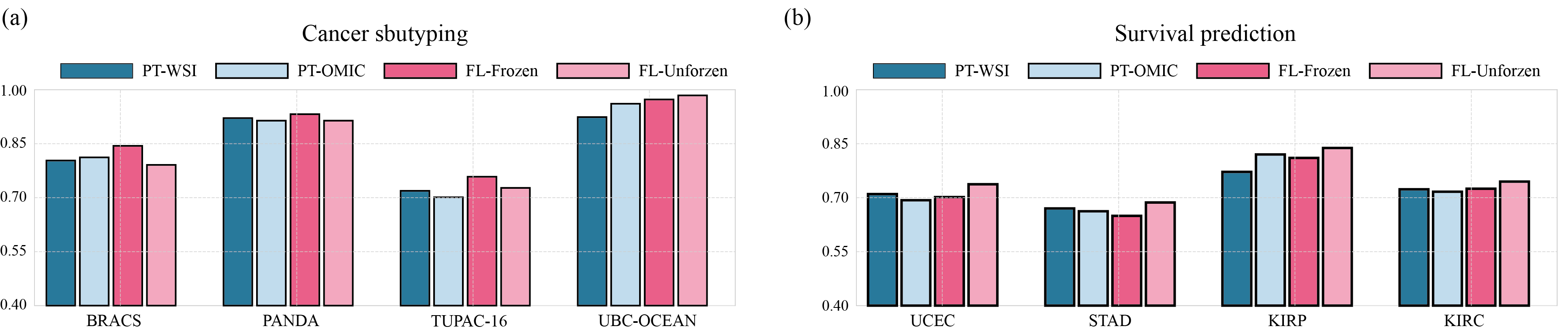}
\caption{Analysis of the performance of ALTER. For cancer subtyping, a unimodal task, we freeze the fusion blocks to prevent multimodal information from being corrupted; for survival prediction as a multimodal task, we unfreeze them, where PT stands for pretraining, and FL stands for fusion layer.} \label{abla}
\vspace{-1em}
\end{figure*}

\begin{figure*}[t]
\centering
\includegraphics[width=0.97\textwidth]{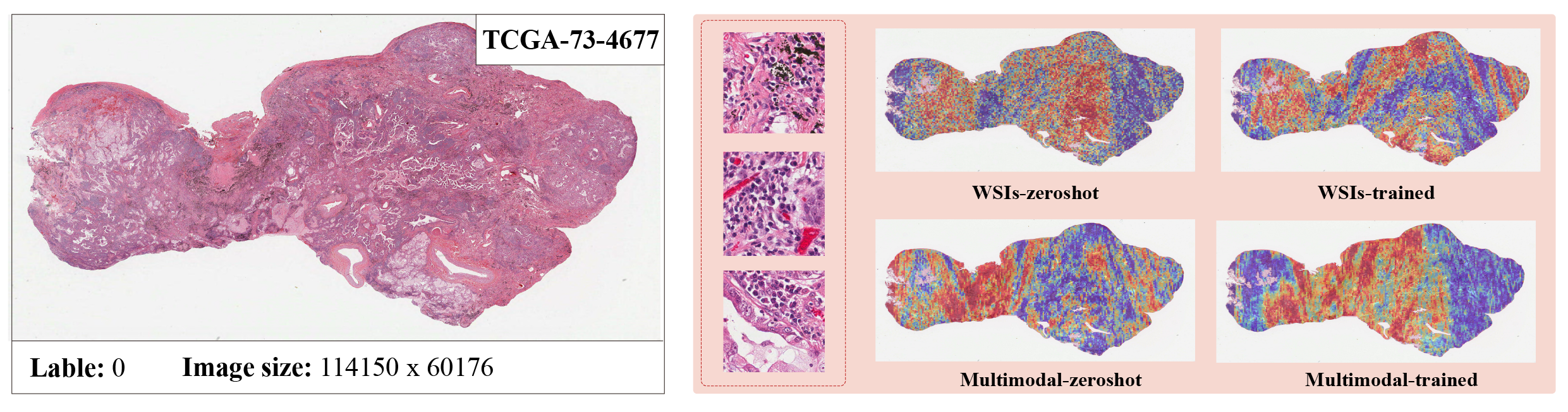}
\caption{Illustration of model’s inspection capabilities on a patient of the LUAD study. The red dashed box highlights the high-risk regions that the trained multimodal model focuses on.} \label{heatmap}
\vspace{-1em}
\end{figure*}

\textbf{Unimodal vs. Multimodal pretraining.} To assess the effect of multimodal input during pretraining, we train unimodal variants of ALTER using either pathology images or omics data alone and evaluate them on various downstream tasks. As shown in Fig.~\ref{abla}, the multimodal model consistently outperforms both unimodal versions, demonstrating the effectiveness of contrastive learning in reducing redundancy and enhancing cross-modal representation. However, this further illustrates that multimodality can enrich information, while also reminding us to consider multi-center data more thoroughly during the pretraining phase, rather than being limited to a single source.
We also examine high-level tissue heatmaps to understand modality-specific attention patterns. As shown in Fig.~\ref{heatmap}, even in zero-shot settings, the multimodal model produces heatmaps closely resembling those of the fully trained model, reflecting strong generalization.

\textbf{Freeze vs. Fine-tune fusion layers.} Another important issue to consider is the impact of whether to fine-tune the fusion layers on model performance. In our experiments, intuitively, for unimodal tasks involving only WSIs, we freeze the fusion layers, while in multimodal survival prediction tasks, we allow them to participate in training. To validate the rationale behind our intuition, we conduct ablation experiments (see Fig.~\ref{abla}). Results show that freezing fusion layers improves performance on unimodal tasks, while fine-tuning yields better results in multimodal settings. This suggests that our pretraining framework successfully preserves inter-modal knowledge within the fusion blocks, which can be effectively leveraged even in unimodal scenarios. Although our analysis focuses on predictive performance, further investigation into the underlying mechanisms—such as attention patterns within the fusion layers—remains an important direction for future work.

\section{Conclusion}
\label{con}

In this article, we propose an any-to-any pretraining framework in the field of computational pathology. Using this framework, we pretrain an any-to-any model and validate its effectiveness through various multimodal downstream tasks. This framework can accept arbitrary modality combinations as input for training, addressing the issue of limited paired data in pathology. We find that with this pretraining framework, even for cancer types that the model has never encountered, fine-tuning can still achieve SOTA performance. While this paper focuses on the TCGA  during the pretraining process, we believe that our framework can be extended to multiple pathology datasets to build a comprehensive foundation model. 
{
    \small
    \bibliographystyle{unsrt}
    \bibliography{main}

\begin{thebibliography}{10}

\bibitem{ma2024towards}
Jiabo Ma, Zhengrui Guo, Fengtao Zhou, Yihui Wang, Yingxue Xu, Yu~Cai, Zhengjie Zhu, Cheng Jin, Yi~Lin~Xinrui Jiang, Anjia Han, et~al.
\newblock Towards a generalizable pathology foundation model via unified knowledge distillation.
\newblock {\em arXiv preprint arXiv:2407.18449}, 2024.

\bibitem{10656058}
Guillaume Jaume, Lukas Oldenburg, Anurag Vaidya, Richard~J. Chen, Drew~F.K. Williamson, Thomas Peeters, Andrew~H. Song, and Faisal Mahmood.
\newblock Transcriptomics-guided slide representation learning in computational pathology.
\newblock In {\em 2024 IEEE/CVF Conference on Computer Vision and Pattern Recognition (CVPR)}, pages 9632--9644, 2024.

\bibitem{xu2024multimodal}
Yingxue Xu, Yihui Wang, Fengtao Zhou, Jiabo Ma, Shu Yang, Huangjing Lin, Xin Wang, Jiguang Wang, Li~Liang, Anjia Han, et~al.
\newblock A multimodal knowledge-enhanced whole-slide pathology foundation model.
\newblock {\em arXiv preprint arXiv:2407.15362}, 2024.

\bibitem{10657501}
Guillaume Jaume, Anurag Vaidya, Richard~J. Chen, Drew~F.K. Williamson, Paul~Pu Liang, and Faisal Mahmood.
\newblock Modeling dense multimodal interactions between biological pathways and histology for survival prediction.
\newblock In {\em 2024 IEEE/CVF Conference on Computer Vision and Pattern Recognition (CVPR)}, pages 11579--11590, 2024.

\bibitem{9710773}
Richard~J. Chen, Ming~Y. Lu, Wei-Hung Weng, Tiffany~Y. Chen, Drew~FK. Williamson, Trevor Manz, Maha Shady, and Faisal Mahmood.
\newblock Multimodal co-attention transformer for survival prediction in gigapixel whole slide images.
\newblock In {\em 2021 IEEE/CVF International Conference on Computer Vision (ICCV)}, pages 3995--4005, 2021.

\bibitem{chen2024towards}
Richard~J Chen, Tong Ding, Ming~Y Lu, Drew~FK Williamson, Guillaume Jaume, Andrew~H Song, Bowen Chen, Andrew Zhang, Daniel Shao, Muhammad Shaban, et~al.
\newblock Towards a general-purpose foundation model for computational pathology.
\newblock {\em Nature Medicine}, 30(3):850--862, 2024.

\bibitem{hemker2023healnet}
Konstantin Hemker, Nikola Simidjievski, and Mateja Jamnik.
\newblock Healnet--hybrid multi-modal fusion for heterogeneous biomedical data.
\newblock {\em arXiv preprint arXiv:2311.09115}, 2023.

\bibitem{araujo2019computing}
Andr{\'e} Araujo, Wade Norris, and Jack Sim.
\newblock Computing receptive fields of convolutional neural networks.
\newblock {\em Distill}, 4(11):e21, 2019.

\bibitem{van2021deep}
Jeroen Van~der Laak, Geert Litjens, and Francesco Ciompi.
\newblock Deep learning in histopathology: the path to the clinic.
\newblock {\em Nature medicine}, 27(5):775--784, 2021.

\bibitem{ilse2018attention}
Maximilian Ilse, Jakub Tomczak, and Max Welling.
\newblock Attention-based deep multiple instance learning.
\newblock In {\em International conference on machine learning}, pages 2127--2136. PMLR, 2018.

\bibitem{campanella2019clinical}
Gabriele Campanella, Matthew~G Hanna, Luke Geneslaw, Allen Miraflor, Vitor Werneck Krauss~Silva, Klaus~J Busam, Edi Brogi, Victor~E Reuter, David~S Klimstra, and Thomas~J Fuchs.
\newblock Clinical-grade computational pathology using weakly supervised deep learning on whole slide images.
\newblock {\em Nature medicine}, 25(8):1301--1309, 2019.

\bibitem{chikontwe2020multiple}
Philip Chikontwe, Meejeong Kim, Soo~Jeong Nam, Heounjeong Go, and Sang~Hyun Park.
\newblock Multiple instance learning with center embeddings for histopathology classification.
\newblock In {\em Medical Image Computing and Computer Assisted Intervention--MICCAI 2020: 23rd International Conference, Lima, Peru, October 4--8, 2020, Proceedings, Part V 23}, pages 519--528. Springer, 2020.

\bibitem{lu2021data}
Ming~Y Lu, Drew~FK Williamson, Tiffany~Y Chen, Richard~J Chen, Matteo Barbieri, and Faisal Mahmood.
\newblock Data-efficient and weakly supervised computational pathology on whole-slide images.
\newblock {\em Nature biomedical engineering}, 5(6):555--570, 2021.

\bibitem{shao2021transmil}
Zhuchen Shao, Hao Bian, Yang Chen, Yifeng Wang, Jian Zhang, Xiangyang Ji, et~al.
\newblock Transmil: Transformer based correlated multiple instance learning for whole slide image classification.
\newblock {\em Advances in neural information processing systems}, 34:2136--2147, 2021.

\bibitem{li2021dt}
Hang Li, Fan Yang, Yu~Zhao, Xiaohan Xing, Jun Zhang, Mingxuan Gao, Junzhou Huang, Liansheng Wang, and Jianhua Yao.
\newblock Dt-mil: deformable transformer for multi-instance learning on histopathological image.
\newblock In {\em Medical Image Computing and Computer Assisted Intervention--MICCAI 2021: 24th International Conference, Strasbourg, France, September 27--October 1, 2021, Proceedings, Part VIII 24}, pages 206--216. Springer, 2021.

\bibitem{xiang2022dsnet}
Tiange Xiang, Yang Song, Chaoyi Zhang, Dongnan Liu, Mei Chen, Fan Zhang, Heng Huang, Lauren O’Donnell, and Weidong Cai.
\newblock Dsnet: A dual-stream framework for weakly-supervised gigapixel pathology image analysis.
\newblock {\em IEEE Transactions on Medical Imaging}, 41(8):2180--2190, 2022.

\bibitem{hou2022h}
Wentai Hou, Lequan Yu, Chengxuan Lin, Helong Huang, Rongshan Yu, Jing Qin, and Liansheng Wang.
\newblock H\^{} 2-mil: exploring hierarchical representation with heterogeneous multiple instance learning for whole slide image analysis.
\newblock In {\em Proceedings of the AAAI conference on artificial intelligence}, volume~36, pages 933--941, 2022.

\bibitem{zheng2022graph}
Yi~Zheng, Rushin~H Gindra, Emily~J Green, Eric~J Burks, Margrit Betke, Jennifer~E Beane, and Vijaya~B Kolachalama.
\newblock A graph-transformer for whole slide image classification.
\newblock {\em IEEE transactions on medical imaging}, 41(11):3003--3015, 2022.

\bibitem{zhang2022dtfd}
Hongrun Zhang, Yanda Meng, Yitian Zhao, Yihong Qiao, Xiaoyun Yang, Sarah~E Coupland, and Yalin Zheng.
\newblock Dtfd-mil: Double-tier feature distillation multiple instance learning for histopathology whole slide image classification.
\newblock In {\em Proceedings of the IEEE/CVF conference on computer vision and pattern recognition}, pages 18802--18812, 2022.

\bibitem{wang2022scl}
Xiyue Wang, Jinxi Xiang, Jun Zhang, Sen Yang, Zhongyi Yang, Ming-Hui Wang, Jing Zhang, Wei Yang, Junzhou Huang, and Xiao Han.
\newblock Scl-wc: Cross-slide contrastive learning for weakly-supervised whole-slide image classification.
\newblock {\em Advances in neural information processing systems}, 35:18009--18021, 2022.

\bibitem{yu2023prototypical}
Jin-Gang Yu, Zihao Wu, Yu~Ming, Shule Deng, Yuanqing Li, Caifeng Ou, Chunjiang He, Baiye Wang, Pusheng Zhang, and Yu~Wang.
\newblock Prototypical multiple instance learning for predicting lymph node metastasis of breast cancer from whole-slide pathological images.
\newblock {\em Medical Image Analysis}, 85:102748, 2023.

\bibitem{lu2023visual}
Ming~Y Lu, Bowen Chen, Andrew Zhang, Drew~FK Williamson, Richard~J Chen, Tong Ding, Long~Phi Le, Yung-Sung Chuang, and Faisal Mahmood.
\newblock Visual language pretrained multiple instance zero-shot transfer for histopathology images.
\newblock In {\em Proceedings of the IEEE/CVF conference on computer vision and pattern recognition}, pages 19764--19775, 2023.

\bibitem{lin2023interventional}
Tiancheng Lin, Zhimiao Yu, Hongyu Hu, Yi~Xu, and Chang-Wen Chen.
\newblock Interventional bag multi-instance learning on whole-slide pathological images.
\newblock In {\em Proceedings of the IEEE/CVF Conference on Computer Vision and Pattern Recognition}, pages 19830--19839, 2023.

\bibitem{li2024dynamic}
Jiawen Li, Yuxuan Chen, Hongbo Chu, Qiehe Sun, Tian Guan, Anjia Han, and Yonghong He.
\newblock Dynamic graph representation with knowledge-aware attention for histopathology whole slide image analysis.
\newblock In {\em Proceedings of the IEEE/CVF Conference on Computer Vision and Pattern Recognition}, pages 11323--11332, 2024.

\bibitem{yang2024mambamil}
Shu Yang, Yihui Wang, and Hao Chen.
\newblock Mambamil: Enhancing long sequence modeling with sequence reordering in computational pathology.
\newblock In {\em International Conference on Medical Image Computing and Computer-Assisted Intervention}, pages 296--306. Springer, 2024.

\bibitem{boehm2022harnessing}
Kevin~M Boehm, Pegah Khosravi, Rami Vanguri, Jianjiong Gao, and Sohrab~P Shah.
\newblock Harnessing multimodal data integration to advance precision oncology.
\newblock {\em Nature Reviews Cancer}, 22(2):114--126, 2022.

\bibitem{ash2021joint}
Jordan~T Ash, Gregory Darnell, Daniel Munro, and Barbara~E Engelhardt.
\newblock Joint analysis of expression levels and histological images identifies genes associated with tissue morphology.
\newblock {\em Nature communications}, 12(1):1609, 2021.

\bibitem{chen2020pathomic}
Richard~J Chen, Ming~Y Lu, Jingwen Wang, Drew~FK Williamson, Scott~J Rodig, Neal~I Lindeman, and Faisal Mahmood.
\newblock Pathomic fusion: an integrated framework for fusing histopathology and genomic features for cancer diagnosis and prognosis.
\newblock {\em IEEE Transactions on Medical Imaging}, 41(4):757--770, 2020.

\bibitem{chen2022pan}
Richard~J Chen, Ming~Y Lu, Drew~FK Williamson, Tiffany~Y Chen, Jana Lipkova, Zahra Noor, Muhammad Shaban, Maha Shady, Mane Williams, Bumjin Joo, et~al.
\newblock Pan-cancer integrative histology-genomic analysis via multimodal deep learning.
\newblock {\em Cancer Cell}, 40(8):865--878, 2022.

\bibitem{jaume2024modeling}
Guillaume Jaume, Anurag Vaidya, Richard~J Chen, Drew~FK Williamson, Paul~Pu Liang, and Faisal Mahmood.
\newblock Modeling dense multimodal interactions between biological pathways and histology for survival prediction.
\newblock In {\em Proceedings of the IEEE/CVF Conference on Computer Vision and Pattern Recognition}, pages 11579--11590, 2024.

\bibitem{li2022hfbsurv}
Ruiqing Li, Xingqi Wu, Ao~Li, and Minghui Wang.
\newblock Hfbsurv: hierarchical multimodal fusion with factorized bilinear models for cancer survival prediction.
\newblock {\em Bioinformatics}, 38(9):2587--2594, 2022.

\bibitem{mobadersany2018predicting}
Pooya Mobadersany, Safoora Yousefi, Mohamed Amgad, David~A Gutman, Jill~S Barnholtz-Sloan, Jos{\'e}~E Vel{\'a}zquez~Vega, Daniel~J Brat, and Lee~AD Cooper.
\newblock Predicting cancer outcomes from histology and genomics using convolutional networks.
\newblock {\em Proceedings of the National Academy of Sciences}, 115(13):E2970--E2979, 2018.

\bibitem{schmauch2020deep}
Beno{\^\i}t Schmauch, Alberto Romagnoni, Elodie Pronier, Charlie Saillard, Pascale Maill{\'e}, Julien Calderaro, Aur{\'e}lie Kamoun, Meriem Sefta, Sylvain Toldo, Mikhail Zaslavskiy, et~al.
\newblock A deep learning model to predict rna-seq expression of tumours from whole slide images.
\newblock {\em Nature communications}, 11(1):3877, 2020.

\bibitem{xie2024spatially}
Ronald Xie, Kuan Pang, Sai Chung, Catia Perciani, Sonya MacParland, Bo~Wang, and Gary Bader.
\newblock Spatially resolved gene expression prediction from histology images via bi-modal contrastive learning.
\newblock {\em Advances in Neural Information Processing Systems}, 36, 2024.

\bibitem{xu2023multimodal}
Yingxue Xu and Hao Chen.
\newblock Multimodal optimal transport-based co-attention transformer with global structure consistency for survival prediction.
\newblock In {\em Proceedings of the IEEE/CVF International Conference on Computer Vision}, pages 21241--21251, 2023.

\bibitem{zhou2023cross}
Fengtao Zhou and Hao Chen.
\newblock Cross-modal translation and alignment for survival analysis.
\newblock In {\em Proceedings of the IEEE/CVF International Conference on Computer Vision}, pages 21485--21494, 2023.

\bibitem{song2024multimodal}
Andrew~H. Song, Richard~J. Chen, Guillaume Jaume, Anurag~Jayant Vaidya, Alexander Baras, and Faisal Mahmood.
\newblock Multimodal prototyping for cancer survival prediction.
\newblock In Ruslan Salakhutdinov, Zico Kolter, Katherine Heller, Adrian Weller, Nuria Oliver, Jonathan Scarlett, and Felix Berkenkamp, editors, {\em Proceedings of the 41st International Conference on Machine Learning}, volume 235 of {\em Proceedings of Machine Learning Research}, pages 46050--46073. PMLR, 21--27 Jul 2024.

\bibitem{meng2024genomics}
Fangliangzi Meng, Hongrun Zhang, Ruodan Yan, Guohui Chuai, Chao Li, and Qi~Liu.
\newblock Genomics-guided representation learning for pathologic pan-cancer tumor microenvironment subtype prediction.
\newblock In {\em International Conference on Medical Image Computing and Computer-Assisted Intervention}, pages 206--216. Springer, 2024.

\bibitem{hemker2024healnet}
Konstantin Hemker, Nikola Simidjievski, and Mateja Jamnik.
\newblock Healnet: Multimodal fusion for heterogeneous biomedical data.
\newblock {\em Advances in Neural Information Processing Systems}, 37:64479--64498, 2024.

\bibitem{guo2024histgen}
Zhengrui Guo, Jiabo Ma, Yingxue Xu, Yihui Wang, Liansheng Wang, and Hao Chen.
\newblock Histgen: Histopathology report generation via local-global feature encoding and cross-modal context interaction.
\newblock In {\em International Conference on Medical Image Computing and Computer-Assisted Intervention}, pages 189--199. Springer, 2024.

\bibitem{chen2022scaling}
Richard~J Chen, Chengkuan Chen, Yicong Li, Tiffany~Y Chen, Andrew~D Trister, Rahul~G Krishnan, and Faisal Mahmood.
\newblock Scaling vision transformers to gigapixel images via hierarchical self-supervised learning.
\newblock In {\em Proceedings of the IEEE/CVF Conference on Computer Vision and Pattern Recognition}, pages 16144--16155, 2022.

\bibitem{lazard2023giga}
Tristan Lazard, Marvin Lerousseau, Etienne Decenci{\`e}re, and Thomas Walter.
\newblock Giga-ssl: Self-supervised learning for gigapixel images.
\newblock In {\em Proceedings of the IEEE/CVF Conference on Computer Vision and Pattern Recognition}, pages 4305--4314, 2023.

\bibitem{yu2023slpd}
Zhimiao Yu, Tiancheng Lin, and Yi~Xu.
\newblock Slpd: slide-level prototypical distillation for wsis.
\newblock In {\em International conference on medical image computing and computer-assisted intervention}, pages 259--269. Springer, 2023.

\bibitem{wu2023position}
Kun Wu, Yushan Zheng, Jun Shi, Fengying Xie, and Zhiguo Jiang.
\newblock Position-aware masked autoencoder for histopathology wsi representation learning.
\newblock In {\em International Conference on Medical Image Computing and Computer-Assisted Intervention}, pages 714--724. Springer, 2023.

\bibitem{jiang2024masked}
Shuai Jiang, Liesbeth Hondelink, Arief~A Suriawinata, and Saeed Hassanpour.
\newblock Masked pre-training of transformers for histology image analysis.
\newblock {\em Journal of Pathology Informatics}, page 100386, 2024.

\bibitem{song2024morphological}
Andrew~H Song, Richard~J Chen, Tong Ding, Drew~FK Williamson, Guillaume Jaume, and Faisal Mahmood.
\newblock Morphological prototyping for unsupervised slide representation learning in computational pathology.
\newblock In {\em Proceedings of the IEEE/CVF Conference on Computer Vision and Pattern Recognition}, pages 11566--11578, 2024.

\bibitem{ding2023pathology}
Kexin Ding, Mu~Zhou, Dimitris~N Metaxas, and Shaoting Zhang.
\newblock Pathology-and-genomics multimodal transformer for survival outcome prediction.
\newblock In {\em International Conference on Medical Image Computing and Computer-Assisted Intervention}, pages 622--631. Springer, 2023.

\bibitem{jin2023gene}
Ting Jin, Xingran Xie, Renjie Wan, Qingli Li, and Yan Wang.
\newblock Gene-induced multimodal pre-training for image-omic classification.
\newblock In {\em International Conference on Medical Image Computing and Computer-Assisted Intervention}, pages 508--517. Springer, 2023.

\bibitem{jaume2024transcriptomics}
Guillaume Jaume, Lukas Oldenburg, Anurag Vaidya, Richard~J Chen, Drew~FK Williamson, Thomas Peeters, Andrew~H Song, and Faisal Mahmood.
\newblock Transcriptomics-guided slide representation learning in computational pathology.
\newblock In {\em Proceedings of the IEEE/CVF Conference on Computer Vision and Pattern Recognition}, pages 9632--9644, 2024.

\bibitem{jaume2024multistain}
Guillaume Jaume, Anurag Vaidya, Andrew Zhang, Andrew~H Song, Richard~J Chen, Sharifa Sahai, Dandan Mo, Emilio Madrigal, Long~Phi Le, and Faisal Mahmood.
\newblock Multistain pretraining for slide representation learning in pathology.
\newblock {\em arXiv preprint arXiv:2408.02859}, 2024.

\bibitem{xu2024whole}
Hanwen Xu, Naoto Usuyama, Jaspreet Bagga, Sheng Zhang, Rajesh Rao, Tristan Naumann, Cliff Wong, Zelalem Gero, Javier Gonz{\'a}lez, Yu~Gu, et~al.
\newblock A whole-slide foundation model for digital pathology from real-world data.
\newblock {\em Nature}, pages 1--8, 2024.

\bibitem{vorontsov2024foundation}
Eugene Vorontsov, Alican Bozkurt, Adam Casson, George Shaikovski, Michal Zelechowski, Kristen Severson, Eric Zimmermann, James Hall, Neil Tenenholtz, Nicolo Fusi, et~al.
\newblock A foundation model for clinical-grade computational pathology and rare cancers detection.
\newblock {\em Nature medicine}, pages 1--12, 2024.

\bibitem{wang2024pathology}
Xiyue Wang, Junhan Zhao, Eliana Marostica, Wei Yuan, Jietian Jin, Jiayu Zhang, Ruijiang Li, Hongping Tang, Kanran Wang, Yu~Li, et~al.
\newblock A pathology foundation model for cancer diagnosis and prognosis prediction.
\newblock {\em Nature}, pages 1--9, 2024.

\bibitem{lu2024visual}
Ming~Y Lu, Bowen Chen, Drew~FK Williamson, Richard~J Chen, Ivy Liang, Tong Ding, Guillaume Jaume, Igor Odintsov, Long~Phi Le, Georg Gerber, et~al.
\newblock A visual-language foundation model for computational pathology.
\newblock {\em Nature Medicine}, 30(3):863--874, 2024.

\bibitem{huang2023visual}
Zhi Huang, Federico Bianchi, Mert Yuksekgonul, Thomas~J Montine, and James Zou.
\newblock A visual--language foundation model for pathology image analysis using medical twitter.
\newblock {\em Nature medicine}, 29(9):2307--2316, 2023.

\bibitem{vaswani2017attention}
Ashish Vaswani, Noam Shazeer, Niki Parmar, Jakob Uszkoreit, Llion Jones, Aidan~N Gomez, {\L}ukasz Kaiser, and Illia Polosukhin.
\newblock Attention is all you need.
\newblock {\em Advances in neural information processing systems}, 30, 2017.

\bibitem{yang2022scbert}
Fan Yang, Wenchuan Wang, Fang Wang, Yuan Fang, Duyu Tang, Junzhou Huang, Hui Lu, and Jianhua Yao.
\newblock scbert as a large-scale pretrained deep language model for cell type annotation of single-cell rna-seq data.
\newblock {\em Nature Machine Intelligence}, 4(10):852--866, 2022.

\bibitem{lee2020biobert}
Jinhyuk Lee, Wonjin Yoon, Sungdong Kim, Donghyeon Kim, Sunkyu Kim, Chan~Ho So, and Jaewoo Kang.
\newblock Biobert: a pre-trained biomedical language representation model for biomedical text mining.
\newblock {\em Bioinformatics}, 36(4):1234--1240, 2020.

\bibitem{radford2021learning}
Alec Radford, Jong~Wook Kim, Chris Hallacy, Aditya Ramesh, Gabriel Goh, Sandhini Agarwal, Girish Sastry, Amanda Askell, Pamela Mishkin, Jack Clark, et~al.
\newblock Learning transferable visual models from natural language supervision.
\newblock In {\em International conference on machine learning}, pages 8748--8763. PmLR, 2021.

\bibitem{klambauer2017self}
G{\"u}nter Klambauer, Thomas Unterthiner, Andreas Mayr, and Sepp Hochreiter.
\newblock Self-normalizing neural networks.
\newblock {\em Advances in neural information processing systems}, 30, 2017.

\bibitem{zhang2024prototypical}
Yilan Zhang, Yingxue Xu, Jianqi Chen, Fengying Xie, and Hao Chen.
\newblock Prototypical information bottlenecking and disentangling for multimodal cancer survival prediction.
\newblock {\em arXiv preprint arXiv:2401.01646}, 2024.

\bibitem{li2021dual}
Bin Li, Yin Li, and Kevin~W Eliceiri.
\newblock Dual-stream multiple instance learning network for whole slide image classification with self-supervised contrastive learning.
\newblock In {\em Proceedings of the IEEE/CVF conference on computer vision and pattern recognition}, pages 14318--14328, 2021.

\bibitem{brancati2022bracs}
Nadia Brancati, Anna~Maria Anniciello, Pushpak Pati, Daniel Riccio, Giosu{\`e} Scognamiglio, Guillaume Jaume, Giuseppe De~Pietro, Maurizio Di~Bonito, Antonio Foncubierta, Gerardo Botti, et~al.
\newblock Bracs: A dataset for breast carcinoma subtyping in h\&e histology images.
\newblock {\em Database}, 2022:baac093, 2022.

\bibitem{bulten2022artificial}
Wouter Bulten, Kimmo Kartasalo, Po-Hsuan~Cameron Chen, Peter Str{\"o}m, Hans Pinckaers, Kunal Nagpal, Yuannan Cai, David~F Steiner, Hester Van~Boven, Robert Vink, et~al.
\newblock Artificial intelligence for diagnosis and gleason grading of prostate cancer: the panda challenge.
\newblock {\em Nature medicine}, 28(1):154--163, 2022.

\bibitem{veta2019predicting}
Mitko Veta, Yujing~J Heng, Nikolas Stathonikos, Babak~Ehteshami Bejnordi, Francisco Beca, Thomas Wollmann, Karl Rohr, Manan~A Shah, Dayong Wang, Mikael Rousson, et~al.
\newblock Predicting breast tumor proliferation from whole-slide images: the tupac16 challenge.
\newblock {\em Medical image analysis}, 54:111--121, 2019.

\bibitem{asadi2024machine}
Maryam Asadi-Aghbolaghi, Hossein Farahani, Allen Zhang, Ardalan Akbari, Sirim Kim, Ashley Chow, Sohier Dane, OCEAN~Challenge Consortium, OTTA Consortium, David~G Huntsman, et~al.
\newblock Machine learning-driven histotype diagnosis of ovarian carcinoma: insights from the ocean ai challenge.
\newblock {\em medRxiv}, pages 2024--04, 2024.

\bibitem{tsuneki2022inferencecaptionshistopathologicalpatches}
Masayuki Tsuneki and Fahdi Kanavati.
\newblock Inference of captions from histopathological patches, 2022.

\bibitem{chen2024wsicaption}
Pingyi Chen, Honglin Li, Chenglu Zhu, Sunyi Zheng, Zhongyi Shui, and Lin Yang.
\newblock Wsicaption: Multiple instance generation of pathology reports for gigapixel whole-slide images.
\newblock In {\em International Conference on Medical Image Computing and Computer-Assisted Intervention}, pages 546--556. Springer, 2024.

\bibitem{du2019gene2vec}
Jingcheng Du, Peilin Jia, Yulin Dai, Cui Tao, Zhongming Zhao, and Degui Zhi.
\newblock Gene2vec: distributed representation of genes based on co-expression.
\newblock {\em BMC genomics}, 20:7--15, 2019.

\bibitem{tomczak2015review}
Katarzyna Tomczak, Patrycja Czerwi{\'n}ska, and Maciej Wiznerowicz.
\newblock Review the cancer genome atlas (tcga): an immeasurable source of knowledge.
\newblock {\em Contemporary Oncology/Wsp{\'o}{\l}czesna Onkologia}, 2015(1):68--77, 2015.

\bibitem{li2023integrative}
He~Li, Lei Yang, Yuanyuan Wang, Lingchan Wang, Gang Chen, Li~Zhang, and Dongchang Wang.
\newblock Integrative analysis of tp53 mutations in lung adenocarcinoma for immunotherapies and prognosis.
\newblock {\em BMC bioinformatics}, 24(1):155, 2023.

\bibitem{lopiccolo2024lung}
Jaclyn LoPiccolo, Alexander Gusev, David~C Christiani, and Pasi~A J{\"a}nne.
\newblock Lung cancer in patients who have never smoked—an emerging disease.
\newblock {\em Nature Reviews Clinical Oncology}, 21(2):121--146, 2024.

\bibitem{zadeh2020bias}
Shekoufeh~Gorgi Zadeh and Matthias Schmid.
\newblock Bias in cross-entropy-based training of deep survival networks.
\newblock {\em IEEE transactions on pattern analysis and machine intelligence}, 43(9):3126--3137, 2020.

\bibitem{wong1986theory}
Wing~Hung Wong.
\newblock Theory of partial likelihood.
\newblock {\em The Annals of statistics}, pages 88--123, 1986.

\end{thebibliography}
}

\newpage
\appendix

\section{Algorithm Pseudo Code of ALTER} \label{pseudo}

\begin{algorithm}[h]
\caption{ALTER: Any-to-Any Learning  via
Triplet Multimodal Pretraining for CPath}
\label{alg:alter}
\begin{algorithmic}[1]
\STATE \textbf{Input:} Dataset $\mathcal{D} = \{(x^{h}, x^{g}, x^{t})\}$ with optional WSI ($x^h$), omics ($x^g$), and reports ($x^t$)
\STATE \textbf{Initialize:} Modality-specific encoders $\mathcal{E}_h, \mathcal{E}_g, \mathcal{E}_t$, Universal Sequence Transformer $\mathcal{U}$

\FOR{epoch $= 1$ to $N_{\text{pretrain}}$}
    \FOR{each batch in $\mathcal{D}$}
        \STATE // Encode available modalities:
        \IF{$x^h$ is available} \STATE $z^h \leftarrow \mathcal{E}_h(x^h)$ \ENDIF
        \IF{$x^g$ is available} \STATE $z^g \leftarrow \mathcal{E}_g(x^g)$ \ENDIF
        \IF{$x^t$ is available} \STATE $z^t \leftarrow \mathcal{E}_t(x^t)$ \ENDIF
        \STATE $Z \leftarrow \text{Concat}(\{z^h, z^g, z^t\})$
        \STATE $F \leftarrow \mathcal{U}(Z)$
        \STATE // Compute losses:
        \IF{epoch  $==0~(mod ~10)$} \STATE $c \leftarrow \text{Random}(h, g, t)$    \quad // Random choose one modality to perform MLM \ENDIF
        \STATE $\mathcal{L}_{CLIP} \leftarrow \text{CLIP}(F^h, F^g, F^t)$ \quad // Equation 9
        \STATE $\mathcal{L}_{triplelet} \leftarrow \text{TripleletLoss}(F)$ \quad // Equation 10
        \STATE $\mathcal{L}_{MLM} \leftarrow \text{MLM}_{c}(F)$  \quad // Modality-specific MLM of the chosen modality $c$
        \STATE $\mathcal{L} \leftarrow \mathcal{L}_{CLIP} + \mathcal{L}_{triplelet} + \mathcal{L}_{MLM}$
        \STATE Update $(\mathcal{E}_w, \mathcal{E}_o, \mathcal{E}_r, \mathcal{U})$ to minimize $\mathcal{L}$
    \ENDFOR
\ENDFOR
\end{algorithmic}
\end{algorithm}

\begin{algorithm}[h]
\caption{Universal Sequence Transformer Module}
\label{alg:universal}
\begin{algorithmic}[1]
\STATE \textbf{Input:} Multi-modal features $Z = \{z^h, z^g, z^t\}$ from encoders
\STATE \textbf{Initialize:} Modality-specific experts $E_h, E_g, E_t$, shared multi-head self-attention (MSA)
\STATE  \textbf{Output:} Shared representation $F$
\STATE
\STATE // Modality-Shared Fusion 
\STATE $Z' \leftarrow \text{ModalitySharedMSA}(Z) + Z$
\STATE // Modality-Specific Decoupling 
\IF{$z^{h'}$ exists} \STATE $z^{h''} \leftarrow E_{h}(z^{h'})$ \ENDIF
\IF{$z^{g'}$ exists} \STATE $z^{g''} \leftarrow E_{g}(z^{g'})$ \ENDIF
\IF{$z^{t'}$ exists} \STATE $z^{t''} \leftarrow E_{t}(z^{t'})$ \ENDIF

\STATE $Z'' \leftarrow \text{concat}(\{z^{h''}, z^{g''}, z^{t''}\})$
\STATE $F \leftarrow Z'' + Z'$
\RETURN  $F$
\end{algorithmic}
\end{algorithm}

\newpage

\section{Data Processing}

\textbf{Whole Slide Images.}  Building on previous works \cite{lu2021data}, we decompose the informative regions of each WSI into a set of non-overlapping $256 \times 256$ pixels patches at $20\times$-equivalent. This way, each WSI can be formulated as multiple instances bag, $W_{i} = \{ w_{1}, w_{2}, \cdots, w_{N_{h}} \}$, where $w_{j}$ is a patch of the WSI and $N_{h}$ is the number of patches.
Then we adopt a pretrained patch encoder UNI \cite{chen2024towards}, a model pretrained with more than 100 million images  across 20 major tissue types, denoted as $f_{w}$ to extract patch-level features. After that, we can obtain a set of features for each WSI, such that $H_{i} = f_{w}(W_{i}) \in \mathbb{R}^{N_{h} \times d_{w}}$, where $d_w = 1024$.

\textbf{Genomic Profiles.}  For genomic profiles, we access RNA sequencing data (RNA-Seq) in this study, which is one of the most widely used  omics data.  Following previous studies \cite{xu2024multimodal},  we employ Gene2Vec \cite{du2019gene2vec}  to provide better embedding representations for gene names. We preserve genes from the Gene2Vec vocabulary to filter out a set of genes applicable to various data sources, such that $G_{i} \in \mathbb{R}^{N_{g}}$, where $N_g$ is the number of genes after selection.

\textbf{Diagnostic Reports.}  Building on previous works \cite{chen2024wsicaption}, we obtain public text from TCGA. Then we use Optical Character Recognition tools to convert them from PDF to text format. After that, we use a tokenizer  to map the text into tokens. Considering that the majority of texts have word counts below 500, we truncate tokens exceeding 512 and pad those below 512. In the end, we obtain $T_{i} \in \mathbb{R}^{N_{t}}$, where $N_{t} = 512$.

\section{Datasets}
\textbf{TCGA Subsets\footnote{\url{https://portal.gdc.cancer.gov/}} \cite{tomczak2015review} for survival prediction.}

We use four TCGA sub-datasets for unimodal survival analysis, each containing WSIs and gene expressions with corresponding survival outcome data:
\begin{enumerate}
    \item TCGA-UCEC: Comprises 480 WSI-omic pairs of uterine corpus endometrial carcinoma, representing various histological grades and disease stages. The dataset provides insights into morphological diversity in endometrial cancer and its prognostic implications.
    \item TCGA-STAD: Includes 317 WSI-omic pairs of stomach adenocarcinoma, covering multiple anatomical regions and histological subtypes of gastric cancer. The associated survival data enables investigation into morphology-outcome relationships.
    \item TCGA-KIRP: Comprises 284 WSI-omic pairs of kidney renal papillary cell carcinoma, representing a distinct subtype with papillary architecture and unique cellular patterns. The dataset facilitates comparative survival modeling across renal cancer subtypes.
    \item TCGA-KIRC: Contains 218 WSI-omic pairs of kidney renal clear cell carcinoma, characterized by clear cytoplasm and variable tumor grades. This dataset enables investigation into survival-relevant morphological features in renal cancers.
\end{enumerate}

All TCGA datasets were digitized under standardized protocols and scanned at 40× magnification, providing high-resolution visualization of cellular and tissue structures. Each dataset is accompanied by rich clinical metadata, including survival outcomes, enabling the development and evaluation of computational models for prognostication based on histopathological features. These comprehensive resources support research in precision oncology and the advancement of personalized medicine.

\textbf{BRACS \footnote{\url{https://www.bracs.icar.cnr.it/}} \cite{brancati2022bracs} for breast cancer subtyping.}

BRACS comprises 547 WSIs and 4,539 regions of interest extracted from those slides. Each WSI and corresponding ROI has been annotated based on the consensus of three board-certified pathologists, ensuring high-quality and reliable labeling. The dataset covers three primary lesion categories—benign, malignant, and atypical—which are further subtyped into seven diagnostic classes: Normal, Pathological Benign, Usual Ductal Hyperplasia, Flat Epithelial Atypia, Atypical Ductal Hyperplasia , Ductal Carcinoma In Situ, and Invasive Carcinoma.

\textbf{PANDA \footnote{\url{https://panda.grand-challenge.org/data/}} \cite{bulten2022artificial} for prostate cancer subtyping.}

PANDA consists of 10,616 core needle biopsy whole-slide images (WSIs) for prostate cancer grading based on the International Society of Urological Pathology (ISUP) system. Following tissue segmentation and the exclusion of slides with low tumor content, a total of 10,202 WSIs were retained for analysis. Each slide is annotated according to one of six ISUP grades, reflecting increasing levels of tumor aggressiveness. This large-scale dataset serves as a benchmark for developing and validating automated methods for histopathological grading of prostate cancer, a critical factor in clinical decision-making and treatment planning.

\textbf{TUPAC-16 \footnote{\url{https://tupac.grand-challenge.org/}} \cite{veta2019predicting} for breast cancer subtyping.}

TUPAC-16 provides a benchmark dataset for predicting tumor proliferation scores from breast cancer whole-slide images. he dataset includes 500 WSIs for training and 321 WSIs for testing, all derived from breast cancer histopathology samples. This dataset supports the development of computational methods aimed at automating tumor proliferation assessment, a key prognostic factor in breast cancer management.

\textbf{UBC-OCEAN \footnote{\url{https://www.kaggle.com/competitions/UBC-OCEAN}} \cite{asadi2024machine} for ovarian cancer subtyping.}

The UBC-OCEAN dataset is a large-scale collection of 527 ovarian cancer WSIs, aggregated from over 20 medical centers spanning four continents. It includes five major histological subtypes—high-grade serous carcinoma, clear-cell carcinoma, endometrioid, low-grade serous, and mucinous carcinoma—as well as several rare subtypes. These subtypes exhibit distinct morphological, molecular, and clinical profiles, making accurate classification essential for effective treatment planning. The challenge of precise subtyping is further compounded by the global shortage of gynecologic pathologists and the increasing demand for subtype-specific therapeutic strategies.

\textbf{TCGA-LUAD \footnote{\url{https://portal.gdc.cancer.gov/}} \cite{tomczak2015review} for gene mutation prediction.}

The TCGA-LUAD dataset comprises 469 WSIs of lung adenocarcinoma, annotated with mutation status for key oncogenes and tumor suppressor genes, including TP53 and EGFR. TP53 is one of the most frequently mutated tumor suppressor genes in lung adenocarcinoma, playing a pivotal role in maintaining genomic stability and regulating cancer initiation and progression \cite{li2023integrative}. EGFR mutations, by contrast, are prevalent oncogenic drivers in non-smoking lung adenocarcinoma patients and serve as critical biomarkers for targeted therapies such as tyrosine kinase inhibitors \cite{lopiccolo2024lung}. The dual annotation of TP53 and EGFR mutation status enables the development of deep learning models for genotype prediction directly from histopathological images, supporting non-invasive biomarker discovery and precision oncology.

\textbf{PatchGastricADC22 \footnote{\url{https://zenodo.org/records/6021442}} \cite{tsuneki2022inferencecaptionshistopathologicalpatches} for report generation.}

The PatchGastricADC22 dataset is a large-scale resource designed to support research on automatic diagnostic report generation from H\&E-stained histopathological images. The dataset was constructed by extracting diagnostic captions from pathology reports corresponding to stomach adenocarcinoma endoscopic biopsy specimens and pairing them with image patches derived from the associated WSIs. PatchGastricADC22 encompasses a diverse range of gastric adenocarcinoma subtypes and includes a total of 262,000 annotated image patches. This dataset enables training and evaluation of models for image-to-text generation in computational pathology.

\section{Baselines}
\subsection{Multimodal Methods}
\textbf{MCAT \cite{9710773}.} The Multimodal Co-Attention Transformer (MCAT) is an early fusion framework designed to model interactions between histology and omics data for patient survival prediction. MCAT employs a dense co-attention mechanism to learn cross-modal mappings between histology patches and omic tokens, which are grouped into six biologically functional gene families. The model generates omic-guided histology features through this co-attention mechanism and concatenates them with the original omic representations to produce a fused feature vector for survival estimation. The use of omic prototypes enhances biological interpretability by aligning model attention with known gene family structures.

\textbf{Porpoise \cite{chen2022pan}.} The Pathology-Omic Research Platform for Integrative Survival Estimation (PORPOISE) is a multimodal learning framework designed to integrate whole-slide histopathology images and molecular profiles for patient-level survival prediction. The architecture comprises three primary neural modules. First, an attention-based multiple instance learning  network extracts slide-level representations from WSI patches. This output is subsequently fused with molecular information via downstream modules that enable cross-modal integration and joint modeling of histological and genomic features for survival estimation.

\textbf{MOTCat \cite{xu2023multimodal}.} The Multimodal Optimal Transport-based Co-Attention Transformer (MOTCat) builds upon MCAT by incorporating optimal transport theory to improve token-level alignment between histology and omics. Specifically, MOTCat estimates an optimal transport plan that maps histology patches to gene tokens grouped into six predefined functional families. This optimal alignment is used to select the most informative histology regions, which are then integrated with omic data for downstream prediction. By leveraging transport-based alignment, MOTCat enhances the precision of cross-modal interactions while maintaining biological interpretability.

\textbf{SurvPath \cite{10657501}.}
SurvPath introduces a biologically-informed multimodal framework that extends beyond fixed gene family groupings by incorporating a transcriptomics tokenizer to generate pathway-level representations. These tokens capture high-level cellular functions derived from transcriptomic data and are fused with histology features using a memory-efficient Transformer. The architecture is designed to model intra-pathway interactions as well as interactions between pathway tokens and histology patches, while omitting histology-to-histology dependencies to reduce computational overhead. This design enables SurvPath to achieve a more biologically grounded integration of modalities with improved efficiency.

\textbf{PIBD  \cite{zhang2024prototypical}.}
Prototypical Information Bottlenecking and Disentangling (PIBD) is a multimodal feature selection and decomposition framework designed to reduce both intra-modal and inter-modal redundancy in pathology-omics integration. The framework comprises two core components: the Prototypical Information Bottleneck (PIB) module and the Prototypical Information Disentanglement (PID) module.  This dual-path disentanglement encourages a structured separation of shared and unique signals across modalities, enhancing interpretability and downstream performance.

\subsection{Unimodal Classification Methods}
\textbf{Max/Mean Pooling.}
Mean pooling is one of the most straightforward and widely adopted aggregation strategies in MIL. It treats all instances within a bag equally by computing the average of their feature representations to produce a unified bag-level embedding. Despite its simplicity, mean pooling effectively captures the overall distribution of instance features, making it particularly suitable for scenarios in which all instances contribute valuable information to the final classification.
In contrast, max pooling adopts a more selective mechanism by focusing solely on the most activated or salient features across instances. This approach assumes that the most discriminative instance carries the most informative signal for bag-level prediction. While max pooling can highlight critical patterns, it may overlook the broader distributional characteristics captured by methods like mean pooling, especially in settings where multiple instances contribute collectively to the label.

\textbf{WiKG \cite{li2024dynamic}.}
WiKG presents a dynamic graph-based framework for WSI analysis, leveraging a knowledge graph to model patch-level relationships. The architecture employs a dual-stream embedding strategy: head embeddings capture inter-patch correlations, while tail embeddings quantify each patch's influence on others. Directed edges are dynamically formed between patches based on these embeddings, and a knowledge-aware attention mechanism is used to aggregate information across the graph. By jointly modeling head, tail, and edge embeddings, WiKG captures both long-range dependencies and directional information flow, addressing limitations of conventional instance-based or undirected graph methods in histopathological analysis.

\textbf{ABMIL \cite{ilse2018attention}.}
Attention-Based Multiple Instance Learning (ABMIL) enhances feature aggregation by introducing a learnable attention mechanism that adaptively assigns importance scores to instances within a bag. Unlike fixed pooling methods, ABMIL computes attention weights based on instance-specific representations, allowing the model to emphasize informative instances and suppress less relevant ones. This adaptive weighting not only improves performance in heterogeneous bags but also provides interpretability, as the learned attention weights reveal the relative contribution of each instance to the classification outcome.

\textbf{DS-MIL \cite{li2021dual}.}
Dual-Stream Multiple Instance Learning (DS-MIL) introduces a two-stream architecture that integrates instance-level and bag-level learning. The first stream uses max pooling to identify a critical instance—the one deemed most discriminative—while the second stream computes attention scores for all instances based on their learned distances to this critical instance. By jointly modeling these two perspectives, DS-MIL captures both localized discriminative features and the broader contextual relevance of other instances. The inclusion of trainable distance metrics further refines instance interactions, making the model well-suited for tasks with complex intra-bag variability.

\textbf{TransMIL \cite{shao2021transmil}.}
TransMIL introduces a correlated MIL framework that departs from the assumption of independent and identically distributed instances. By leveraging a Transformer-based architecture, TransMIL captures both morphological and spatial correlations among instances in WSIs. The model incorporates a Transformer Pyramid Translation (TPT) module that combines multi-head self-attention with position-aware encoding to preserve spatial context. This design allows the model to effectively integrate relational information across patches, improving classification performance on complex histopathological data.

\textbf{CLAM-SB \cite{lu2021data}.}
The Clustering-constrained Attention Multiple Instance Learning (CLAM) framework is designed for weakly supervised WSI classification. In CLAM, image patches are first encoded into fixed feature representations using a pre-trained CNN. During both training and inference, these feature vectors serve as inputs to an attention-based multiple instance learning model. The attention mechanism aggregates patch-level features into a slide-level representation, which is then used for final diagnostic prediction. This design enables CLAM to efficiently learn from slide-level labels without requiring detailed patch annotations, while providing interpretability through the attention scores assigned to individual patches.

\subsection{Unimodal Generation Methods}
\textbf{WSICaption \cite{chen2024wsicaption}.}
WsiCaption introduces a novel framework for generating pathology reports directly from WSIs, facilitating multimodal learning in computational pathology. To enable effective learning from WSIs, WsiCaption employs a multiple instance generation framework (MI-Gen) that treats each WSI as a bag of image patches. A position-aware module is integrated into the model to enhance sensitivity to spatial context, allowing the model to generate more coherent and spatially informed textual descriptions. This design improves the interpretability and clinical relevance of the generated pathology reports.

\textbf{HistGen \cite{guo2024histgen}.}
HistGen is a hierarchical framework for efficient and scalable WSI representation learning. It adopts a region-to-slide paradigm, wherein each WSI is first divided into regions that are individually processed using a pre-trained vision transformer. Subsequently, a series of attention-based modules hierarchically aggregate the regional features to produce slide-level embeddings. This local-global encoding strategy allows the model to preserve meaningful spatial relationships while managing the computational demands posed by gigapixel images. Notably, HistGen demonstrates strong transferability and robustness by bridging local tissue features and global contextual patterns.

\section{Task-specific Losses}

\subsection{Survival Loss Functions}
Survival analysis models the time until an event occurs, which may not always be observed (i.e., it can be censored). In cancer survival prediction, a censored event indicates either patient survival or the last known follow-up time, while an uncensored event signifies patient death. We now detail the Negative log-likelihood and Cox proportional Hazards survival losses.

\subsubsection{Negative Log-Likelihood Loss}
For survival prediction, we employ a negative log-likelihood (NLL) loss that is suitable for right-censored data, following prior work \cite{zadeh2020bias,song2024multimodal}. The objective is to model the survival outcome using a patient-level embedding $\bar{x}_{\text{patient}} \in \mathbb{R}^{2d}$. Each patient sample includes: (1) a censorship indicator $c \in {0, 1}$, where $c = 0$ indicates an observed event (death) and $c = 1$ denotes a censored instance (last known follow-up), and (2) a time-to-event value $t_i$.

Instead of regressing directly on $t_i$, we discretize the time axis into $n$ non-overlapping intervals $(t_{j-1}, t_j]$, determined by the quartiles of observed survival times. Each interval is represented by a categorical label $y_j$, transforming the survival prediction task into a classification problem over discrete time bins. For each interval, we define a hazard function, $f_{\text{hazard}}(y_j \mid \bar{x}_{\text{patient}}) = \sigma(\hat{y}_j)$, where $\sigma$ is the sigmoid function, and $f_{\text{hazard}}$ denotes the probability of death within interval $(t_{j-1}, t_j]$. The corresponding survival function is given by:

\begin{equation}
f_{\text{surv}}(y_j|\bar{x}_{\text{patient}}) = \prod_{k=1}^j (1 - f_{\text{hazard}}(y_k|\bar{x}_{\text{patient}})),
\end{equation}

which computes the probability of surviving up to the end of interval $t_j$. The overall NLL survival loss over a dataset of $N_D$ patients is defined as:

\begin{align}
\mathcal{L}&\left(\{\bar{x}_{\text{patient}}^{(i)}, y_j^{(i)}, c^{(i)}\}_{i=1}^{N_D}\right) = \sum_{i=1}^{N_D} -c^{(i)} \log(f_{\text{surv}}(y_j^{(i)}|\bar{x}_{\text{patient}}^{(i)})) \nonumber \\
&+ (1-c^{(i)}) \log(f_{\text{surv}}(y_j^{(i)}-1|\bar{x}_{\text{patient}}^{(i)})) + (1-c^{(i)}) \log(f_{\text{hazard}}(y_j^{(i)}|\bar{x}_{\text{patient}}^{(i)}))
\end{align}

The first term encourages high predicted survival probabilities for censored patients up to their last follow-up time. The second term enforces survival probability up to the interval preceding death in uncensored cases. The third term encourages accurate identification of the specific interval in which death occurs. Together, these components ensure effective modeling of both censored and uncensored survival data.

\subsubsection{Cox Proportional Hazards Loss}
In addition to discrete-time survival modeling, we also consider the Cox proportional hazards model, a widely used method for analyzing right-censored survival data. The model parameterizes the hazard function $\lambda(t \mid x)$ as:

\begin{equation}
\lambda(t \mid x) = \lambda_0(t) \exp(\theta^\top x),
\end{equation}

where $\lambda_0(t)$ is the baseline hazard function representing the underlying risk of an event over time, and $\theta$ denotes the learnable model parameters. Here, $x$ corresponds to the patient-level embedding $\bar{x}_{\text{patient}} \in \mathbb{R}^{2d}$.

To estimate $\theta$ without requiring specification of $\lambda_0(t)$, the Cox partial log-likelihood is employed \cite{wong1986theory}. This likelihood quantifies the probability of observing events in uncensored patients relative to others at risk:

\begin{equation}
\mathcal{L}\left(\theta, \overline{\mathrm{x}}_{\text {patient }}\right)=-\sum_{i \in U}\left(\overline{\mathrm{x}}_{\text {patient }, i} \theta-\log (\sum_{j \in R_{i}} \exp \left(\overline{\mathrm{x}}_{\text {patient}, j} \theta\right))\right),
\end{equation}

where $\mathcal{U}$ denotes the set of uncensored patients, and $\mathcal{R}_i$ is the risk set for patient $i$, consisting of all patients still under observation (i.e., uncensored or censored after $i$'s event time). This formulation encourages the model to assign higher relative risk scores to patients who experience events earlier.

The gradient of the partial log-likelihood with respect to the input embeddings is given by:

\begin{equation}
\frac{\partial l\left(\theta, \overline{\mathrm{x}}_{\text {patient }}\right)}{\partial \overline{\mathrm{x}}_{\text {patient }, i}}=\delta(i) \theta-\sum_{i, j \in C_{j}, U} \frac{\theta \exp \left(\overline{\mathrm{x}}_{\text {patient }, i} \theta\right)}{\sum_{k \in C_{j}} \exp \left(\overline{\mathrm{x}}_{\text {patient }, k} \theta\right)}
,
\end{equation}

where $\delta(i)$ is an indicator function that equals 1 if the event for patient $i$ is observed, and $\mathcal{C}$ is the set of censored patients. This gradient formulation enables backpropagation through the Cox loss in a differentiable deep learning pipeline.

\subsection{Other Loss Functions}

For other tasks such as cancer subtyping, gene mutation prediction, and report generation, we adopt the standard cross-entropy (CE) loss. Let a WSI be denoted by $X$ with an associated slide-level label $y$, and let the model produce predictions $\hat{y} = f(X; \theta)$, where $\hat{y} \in \mathbb{R}^C$ represents the predicted probability distribution over $C$ classes. The CE loss for a single WSI is defined as:

\begin{equation}
\mathcal{L}_{\text{ce}} = -\sum_{c=1}^C y_c \log(\hat{y}_c),
\end{equation}

where $y_c$ is the one-hot encoded ground-truth label for class $c$. The overall classification loss is computed by averaging over all $N$ WSIs in the training set:

\begin{equation}
\mathcal{L}_{\text{cls}} = \frac{1}{N} \sum_{i=1}^N \mathcal{L}_{\text{ce}}^{(i)},
\end{equation}

where $\mathcal{L}_{\text{ce}}^{(i)}$ denotes the CE loss for the $i$-th WSI. This objective encourages the model to assign high probabilities to the correct class labels while suppressing incorrect predictions. It also facilitates the learning of a query-aware attention mechanism that identifies diagnostically relevant regions within each WSI.

\section{Implementation Details} \label{implement}
We implement ALTER using PyTorch. For the feature extraction backbone, we utilize CPath pre-trained foundation model UNI \cite{chen2024towards} to obtain 1024-dimensional patch features. The model contains 8 attention heads, with the hidden dimension set to 512.

During the pre-training phase, we train the model using the Adam optimizer with a learning rate of $1e^{-3}$. Considering the requirements of contrastive learning for batch size, we use five A100 GPUs for parallel training and set the batch size to 60. The hyperparameters $\alpha$ and $\beta$ are set to 1. And we randomly select a modality to perform the MLM task after every 10 epochs until the model converged.

For the downstream tasks, we fine-tune the model with details below:

\begin{enumerate}
    \item \textbf{Survival prediction:} We train the model using  the Cox loss, and a batch size of 12. We use a learning rate of $5e^{-4}$ for STAD and $1e^{-4}$ for other datasets.
    \item \textbf{Cancer subtyping:} we train the model with frozen fusion layers using the CE loss, and a batch size of 1. We use a learning rate of $1e^{-5}$ for BRACS and $1e^{-4}$ for other datasets.
    \item \textbf{Gene mutation prediction:} we train the model with frozen fusion layers using the CE loss, and a batch size of 1. We use a learning rate of $1e^{-5}$ for TP53 and $5e^{-5}$ for EGFR.
    \item \textbf{Report generation:} we train the model with frozen fusion layers using the CE loss, a batch size of 1,  and a learning rate of $1e^{-4}$.
\end{enumerate}

All of the models of the downstream tasks are trained with Adam optimizer on a single NVIDIA A100. And all of the baselines are trained with the code reported in the respective papers.

\section{Limitations \& Future Directions} \label{limit}

First, although ALTER is designed to accommodate flexible combinations of input modalities, its current implementation is limited to data from a single source—the TCGA dataset. This constraint primarily arises from the substantial computational resources required for large-scale multimodal pretraining, which restricted our ability to incorporate multi-institutional datasets. Consequently, ALTER has not yet been exposed to the inter-institutional variability characteristic of real-world clinical settings, potentially limiting its robustness and generalizability across diverse clinical environments.

Second, in multimodal tasks such as survival prediction, we observe that unimodal baselines perform inconsistently across different modalities. For example, in the UCEC dataset, models relying solely on omics features demonstrate relatively poor performance, suggesting that some modalities may carry significant noise. While ALTER exhibits robustness to such noise through cross-modal contrastive pretraining, it currently lacks explicit mechanisms to interpret or quantify the individual contributions of each modality. This presents a notable limitation: without improved interpretability, it remains challenging to discern how the model integrates and prioritizes noisy or less-informative modalities during downstream inference.

Finally, our attention map analysis (Fig.~\ref{heatmap}) shows that the multimodal pretrained model exhibits attention patterns closely aligned with those of a fine-tuned model, even in zero-shot scenarios. This contrasts with unimodal pretrained models, which display less task-relevant attention alignment. These observations suggest that multimodal pretraining qualitatively alters the model’s internal representation space. However, the specific influence of the pretraining data composition on attention dynamics remains unclear, underscoring the need for further investigation into how different modality combinations during pretraining affect model interpretability and downstream behavior.


\end{document}